\documentclass[final]{cvpr}
\pdfoutput=1
\usepackage{times}
\usepackage{epsfig}
\usepackage{graphicx}
\usepackage{amsmath}
\usepackage{amssymb}

\usepackage{caption}
\usepackage{algorithm}
\usepackage{algorithmic}
\usepackage{multicol}
\usepackage{multirow}

\usepackage[pagebackref=true,breaklinks=true,colorlinks,bookmarks=false]{hyperref}

\newcommand{\ftst}{\textcolor[rgb]{0,0, 0}}
\newcommand{\ftnd}{\textcolor[rgb]{0.,0,0}}
\newcommand{\jcst}{\textcolor[rgb]{0,0,0}}
\newcommand{\jcnd}{\textcolor[rgb]{0,0.,0}}
\newcommand{\jcgam}{\textcolor[rgb]{0.,0.,0}}
\newcommand{\jcla}{\textcolor[rgb]{0.,0.,0}}
\newcommand{\jcca}{\textcolor[rgb]{0,0,0}}

\renewcommand{\eg}{\textit{e.g.\ }}
\renewcommand{\ie}{\textit{i.e.\ }}
\newcommand{\et}{\textit{et al.\ }}

\def\cvprPaperID{1937}
\def\confYear{CVPR 2021}

\renewenvironment{quote}{
    \onecolumn
}

\begin{document}

\begin{quote}

\newpage
\onecolumn
\noindent \vspace{1cm}
\noindent \textbf{\huge{MIST: Multiple Instance Self-Training Framework for Video Anomaly Detection}}

\vspace{2cm}

\noindent {\LARGE{Jia-Chang Feng, Fa-Ting Hong, Wei-Shi Zheng}}

\vspace{2cm}


\vspace{1cm}

\noindent For reference of this work, please cite:

\vspace{1cm}
\noindent Jia-Chang Feng, Fa-Ting Hong and Wei-Shi Zheng.
``MIST: Multiple Instance Self-Training Framework for Video Anomaly Detection \emph{Proceedings of the IEEE International Conference on Computer Vision and Pattern Recognition.} 2021.

\vspace{1cm}

\noindent Bib:\\
\noindent
@inproceedings\{feng2021mist,\\
\ \ \   title=\{MIST: Multiple Instance Self-Training Framework for Video Anomaly Detection\},\\
\ \ \  author=\{Feng, Jia-Chang and Hong, Fa-Ting and Zheng, Wei-Shi\},\\
\ \ \  booktitle=\{Proceedings of the IEEE International Conference on Computer Vision and Pattern Recognition\},\\
\ \ \  year=\{2021\}\\
\}
\end{quote}

%
\twocolumn
\author{Jia-Chang Feng$^{1,3,4}$, Fa-Ting Hong$^{1,3}$, and Wei-Shi Zheng$^{1,2,3}$\thanks{Corresponding author} \\
{ \small $^1$ School of Computer Science and Engineering, Sun Yat-Sen University} \\
{ \small $^2$ Peng Cheng Laboratory, Shenzhen, China}\\
{ \small $^3$ Key Laboratory of Machine Intelligence and Advanced Computing, Ministry of Education, China}\\
{\small $^4$ Pazhou Lab, Guangzhou, China}\\
{\tt \small fengjch8@mail2.sysu.edu.cn, hongft3@mail2.sysu.edu.cn, wszheng@ieee.org }
}

\title{\jcst{MIST: Multiple Instance Self-Training Framework for Video Anomaly Detection}}

\maketitle

\begin{abstract}

Weakly supervised video anomaly detection (WS-VAD) is to distinguish anomalies from normal events based on discriminative representations. Most existing works are limited in insufficient video representations.
In this work, we develop a multiple instance self-training framework (MIST) to efficiently refine task-specific discriminative representations with only video-level annotations.
In particular, MIST is composed of 1) a multiple instance pseudo label generator, which adapts a sparse continuous sampling strategy to produce more reliable clip-level pseudo labels,
and 2) a self-guided attention boosted feature encoder that aims to automatically focus on anomalous regions in frames while extracting task-specific representations. 
\ftst{Moreover, we adopt a self-training scheme to optimize both components and finally obtain a task-specific feature encoder}.
Extensive experiments on two public datasets demonstrate the efficacy of our method, and our method 
\jcca{performs comparably to or even better than existing supervised and weakly supervised methods, specifically obtaining a frame-level AUC 94.83\% on ShanghaiTech.}
\end{abstract}

\section{Introduction}
\label{ses:intro}

\begin{figure}
    \centering
    \includegraphics[width=\linewidth]{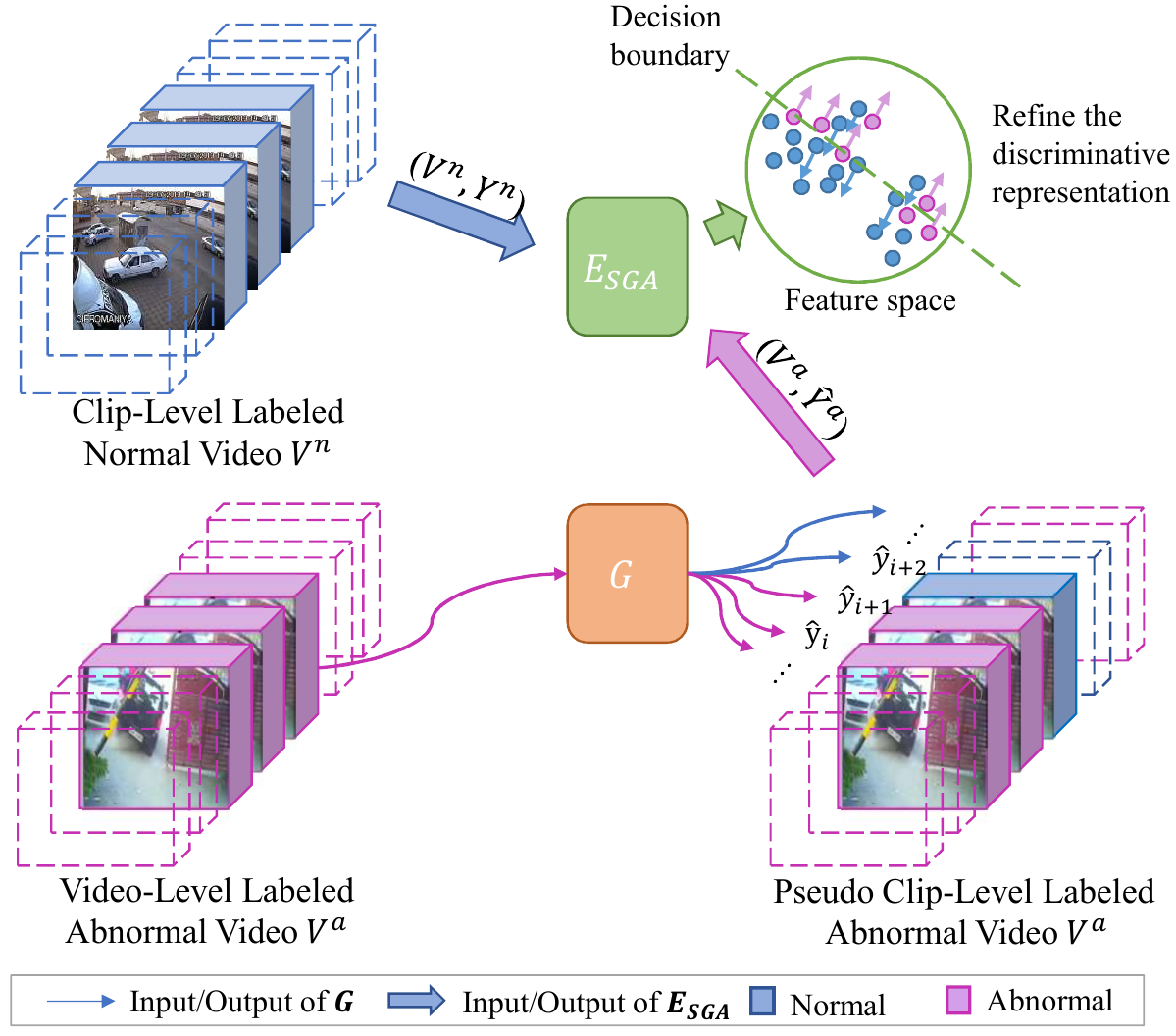}
    \caption{\jcla{Our proposed MIST first assign clip-level pseudo labels $\hat{Y}^{a}=\{\hat{y}^a_i\}$ to anomaly videos with the help of a pseudo label generator $G$. Then, MIST leverages information from all videos to refine a self-guided attention boosted feature encoder $E_{SGA}$.}}
    \label{fig:Front_Update}
\end{figure}

Video anomaly detection (VAD) aims to temporally or spatially localize anomalous events in videos \cite{zhu2020video}. As \jcca{increasingly} more surveillance cameras are deployed, VAD is playing an increasingly important role in intelligent surveillance systems to reduce the manual work of live monitoring.

\jcst{Although VAD has been researched for years,}
developing a model to detect anomalies in videos remains challenging, as it requires the model to understand the inherent differences between normal and abnormal events, especially anomalous events that are rare and vary substantially.
Previous works treat VAD as an \textit{unsupervised learning} task \cite{zhao2011online,lu2013abnormal,hasan2016learning,luo2017revisit,liu2018future,gong2019memorizing,zhou2019anomalynet} \jcst{, which 
\jcgam{ encodes the usual pattern with only normal training samples, and then detects the distinctive encoded patterns as anomalies.}}
Here, we aim to address the \textit{weakly supervised video anomaly detection} (WS-VAD) problem \cite{sultani2018real,zhong2019graph,zhang2019temporal,zhu2019motion,wan2020weakly} \ftst{because obtaining video-level labels is more realistic and can produce more reliable results than unsupervised methods.}
More specifically, existing methods in WS-VAD can be categorized into two classes,
\ie \textit{encoder-agnostic} and \textit{encoder-based} methods.

\jcst{The \textit{encoder-agnostic} methods \jcca{\cite{sultani2018real,zhang2019temporal,wan2020weakly}} utilize task-agnostic features of videos extracted from a vanilla feature encoder denoted as $E$ (\eg C3D \cite{tran2015learning} or I3D \cite{carreira2017quo}) to estimate anomaly scores.
The \textit{encoder-based} methods \jcca{\cite{zhu2019motion,zhong2019graph}} train both the feature encoder and classifier simultaneously.}
\jcnd{The state-of-the-art encoder-based method is
Zhong \et  \cite{zhong2019graph}, which} formulates WS-VAD as a label noise learning problem 
\jcnd{and learns from the noisy labels filtered by a label noise cleaner network.}
However, label noise results from assigning video-level labels to each clip. Even though the cleaner network corrects some of the noisy labels \jcst{in the time-consuming iterative optimization},
\jcst{the refinement of representations progresses slowly }
as \ftst{these models are \jcgam{mistaught} by 
\jcst{seriously noisy pseudo labels}
at the beginning.}


\jcst{We find that the existing methods have not considered training a task-specific feature encoder efficiently, which offers discriminative representations for events under surveillance cameras.}
\ftst{To overcome this problem for WS-VAD, we develop a two-stage self-training procedure (Figure \ref{fig:Front_Update}) that aims to train a task-specific feature encoder with only video-level weak labels.}
In particular, we propose a Multiple Instance Self-Training framework (MIST) that consists of a multiple instance pseudo label generator and a self-guided attention boosted feature encoder $E_{SGA}$.
1) \textit{MIL-pseudo label generator.}
The MIL framework is well verified in weakly supervised learning. MIL-based methods can generate pseudo labels more accurately than those simply assigning video-level labels to each clip \cite{zhong2019graph}. Moreover, we adopt a sparse continuous sampling strategy that can \jcla{force the network to pay more attention to context around the most anomalous part. }
2) \textit{Self-guided attention boosted feature encoder.} Anomalous events in surveillance videos may occur in any place and with any size \cite{liu2019exploring}, while in commonly used action recognition videos, the action usually appears with large motion
\cite{choi2019can,choi2020unsupervised}. Therefore, we utilize the proposed self-guided attention module in our proposed feature encoder to emphasize the anomalous regions without any external annotation \cite{liu2019exploring} but \ftst{clip-level annotations of normal videos and clip-level pseudo labels of anomalous videos. For our WS-VAD modelling, we introduce a deep MIL ranking loss to effectively train the multiple instance pseudo label generator.
\jcca{In particular, for deep MIL ranking loss, we adopt a sparse-continuous sampling strategy to focus more on the context around the anomalous instance.}
}

\ftst{To obtain a task-specific feature encoder \jcst{with smaller domain-gap,}
we introduce an efficient two-stage self-training scheme to optimize the proposed framework. We use the features extracted from \jcst{the} \ftnd{original} feature encoder to produce its corresponding clip-level pseudo labels for anomalous videos by the generator $G$. \jcgam{Then,} we adopt these pseudo labels and their corresponding abnormal videos as well as normal videos to refine \ftnd{our improved} feature encoder $E_{SGA}$ (as demonstrated in Figure \ref{fig:Front_Update}). Therefore, we can acquire a task-specific feature encoder
\jcst{that provides discriminative representations for surveillance videos.}}

\jcgam{The extensive experiments based on two different feature encoders, \ie C3D \cite{tran2015learning} and I3D \cite{carreira2017quo} show that our framework MIST is able to produce a task-specific feature encoder. }
\ftst{We also compare the proposed framework with other encoder-agnostic methods on two large datasets \ie, UCF-Crime \cite{sultani2018real} and ShanghaiTech\cite{luo2017revisit}. In addition, we \jcst{run} ablation studies to evaluate our proposed sparse continuous sampling strategy and self-guided attention module. We also 
illustrate some visualized results to provide a more intuitive understanding of our approach. Our experiments demonstrate the effectiveness and efficiency of MIST.}


\section{Related Works}
\begin{figure*}[ht]
    \centering
    \includegraphics[width=0.95\textwidth]{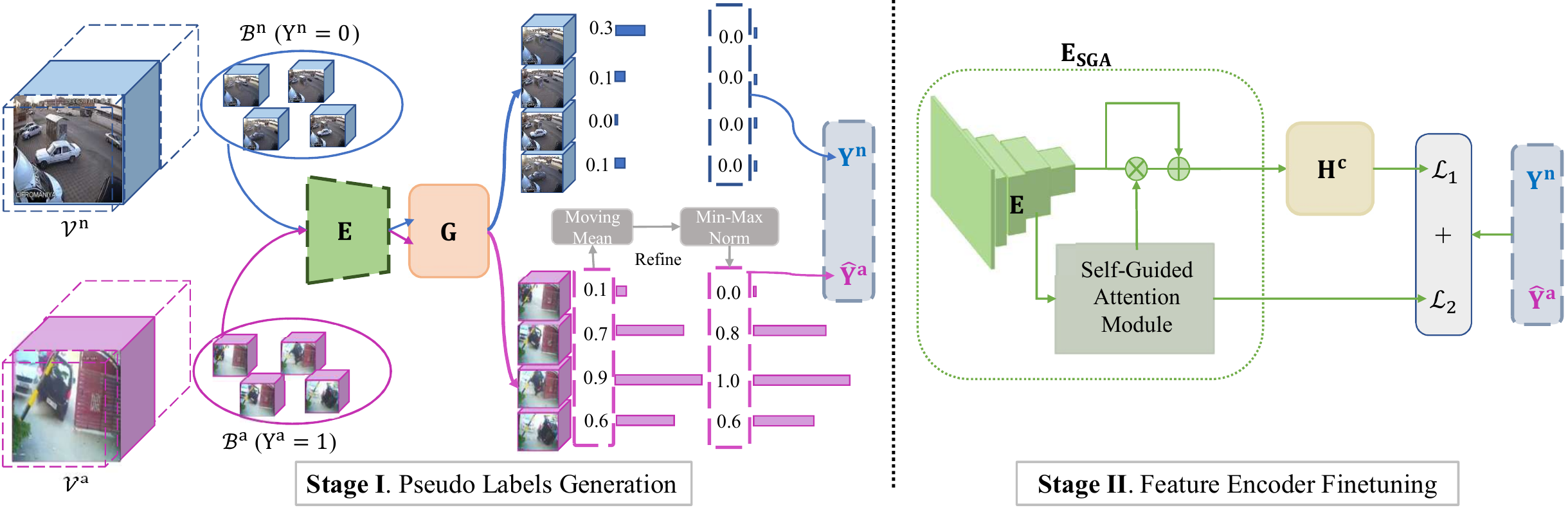}
    \vspace{-0.1cm}
    \caption{Illustration of our proposed MIST framework. MIST includes a multiple instance pseudo label generator $G$ and self-guided attention boosted feature encoder $E_{SGA}$ followed by a weighted-classification head $H_c$. We first train a $G$ and then generate pseudo labels for $E_{SGA}$ fine-tuning.}
    \vspace{-0.1cm}
    \label{fig:Strucutre}
\end{figure*}
\noindent \textbf{Weakly supervised video anomaly detection.} VAD \jcca{aims to detect anomaly events in a given video and} has been researched for years\cite{hospedales2009markov,zhao2011online,lu2013abnormal,hasan2016learning,luo2017revisit,liu2018future,liu2019margin,zhou2019anomalynet,zhong2019graph,gong2019memorizing,wan2020weakly}.
\jcst{Unsupervised learning methods \cite{hospedales2009markov,zhao2011online,hasan2016learning,zhao2017spatio,luo2017revisit,liu2018future,zhou2019anomalynet,gong2019memorizing} }
\jcgam{ encode the usual pattern with only normal training samples and then detect the distinctive encoded patterns as anomalies.}
Weakly supervised learning methods \cite{sultani2018real,zhong2019graph,zhang2019temporal,zhu2019motion,wan2020weakly} \jcnd{with video-level labels} are more applicable to distinguish abnormal events and normal events.
Existing weakly supervised VAD methods can be categorized into two classes, \ie,  \textit{encoder-agnostic} methods and \textit{encoder-based} methods. 1) \textit{Encoder-agnostic} methods train only the classifier. Sultani \et  \cite{sultani2018real} proposed a deep MIL ranking framework to detect anomalies; Zhang \et  \cite{zhang2019temporal} further introduced inner-bag score gap regularization; Wan \et  \cite{wan2020weakly} introduced dynamic MIL loss and center-guided regularization. 
2) \textit{Encoder-based} methods train both a feature encoder and a classifier. Zhu \et \cite{zhu2019motion} proposed an attention based MIL model combined with a optical flow based auto-encoder to encode motion-aware features.
Zhong \et  \cite{zhong2019graph} took weakly supervised VAD as a label noise learning task and proposed GCNs to filter label noise for iterative model training, \jcst{ but the iterative optimization was inefficient and progressed slowly.}
\jcca{Some works focus on detecting anomalies in an offline manner \cite{ullah2020cnn,wu2020not} or a coarse-grained manner \cite{sultani2018real,zhang2019temporal,zhu2019motion,ullah2020cnn,wu2020not}, which do not meet the real-time monitoring requirements for real-world applications.}

Here, our work is also an encoder-based method \jcca{and work in an online fine-grained manner}, but \ftst{we use the learned pseudo labels to optimize our feature encoder $E_{SGA}$ rather than using video-level labels as pseudo labels directly. Moreover, we design a two-stage self-training scheme to efficiently optimize our feature encoder and pseudo label generator instead of iterative optimization\cite{zhong2019graph}.}

\vspace{0.1cm}
\noindent \textbf{Multiple Instance Learning.} MIL is a popular method for weakly supervised learning. \ftnd{In video-related tasks}, MIL takes a video as a bag and clips in the video as instances \cite{sultani2018real,nguyen2018weakly,hong2020mini}. With a specific feature/score aggregation function, video-level labels can be used to indirectly supervise instance-level learning. The aggregation functions vary, \eg max pooling\cite{sultani2018real,zhang2019temporal,zhu2019motion} and attention pooling\cite{nguyen2018weakly,hong2020mini}.
In this paper, we adopt a sparse continuous sampling strategy in our multiple instance pseudo label generator to
\jcnd{ force the network to pay more attention to context around the most anomalous part.}

\vspace{0.1cm}
\noindent \textbf{Self-training.} Self-training has been widely investigated in semi-supervised learning \cite{amini2002semi,lee2013pseudo,grandvalet2005semi,yarowsky1995unsupervised,triguero2015self,zou2019confidence}. 
\jcst{Self-training methods increase labeled data via pseudo label generation on unlabeled data to leverage the information on both labeled and unlabeled data}. \jcst{Recent deep self-training involves representation learning of the feature encoder and classifier refinement, mostly adopted in semi-supervised learning \cite{lee2013pseudo} and domain adaptation \cite{zou2018unsupervised,zou2019confidence}. }
\jcca{In unsupervised VAD, Pang \et \cite{pang2020self} introduced a self-training framework deployed on the testing video directly, assuming the existence of an anomaly in the given video.}

\jcst{Here, we propose a multiple instance self-training framework that assigns clip-level pseudo labels to all clips in abnormal videos via a multiple instance pseudo label generator. Then, we leverage information from all videos to fine-tune a self-guided attention boosted feature encoder. }


\section{Approach}
VAD depends on discriminative representations that clearly represent the events in a scene, while action recognition datasets pretrained feature encoders are not perfect for surveillance videos because of the existence of a domain gap \cite{liu2019exploring,choi2019can,choi2020unsupervised}.
To address this problem, we introduce a self-training strategy to refine the \ftnd{proposed improved} feature encoder $E_{SGA}$. An illustration of our method shown in Figure \ref{fig:Strucutre} is detailed in the following.
\begin{algorithm}[t] 
 \small
 \caption{\jcnd{Multiple instance self-training framework}}
 \label{alg:framework}
 \begin{algorithmic}[1]
    \REQUIRE Clip-level labeled normal videos $V^n = \{v^n_i\}^N_{i=1}$ and corresponding clip-level labels $Y^n$, video-level labeled abnormal videos $V^a = \{v^a_i\}^N_{i=1}$, pretrained vanilla feature encoder $E$.
    \ENSURE Self-guided attention boosted feature encoder $E_{SGA}$, multiple instance pseudo label generator $G$,  clip-level pseudo labels $\hat{Y}^a$ for $V^a$
    
    \noindent\jcst{\textbf{\textit{Stage I. Pseudo Labels Generation.}}}
    \STATE Extract features of $V^a$ and $V^n$ from $E$ as $\{\textup{f}^a_i\}^{N}_{i=1}$ and $\{\textup{f}^n_i\}^{N}_{i=1}$.
    \STATE Training $G$ with $\{\textup{f}^a_i\}^{N}_{i=1}$ and $\{\textup{f}^n_i\}^{N}_{i=1}$ and their corresponding video-level labels according to Eq. \ref{equ:mil_loss}.
    \STATE Predict clip-level pseudo labels for each clip of $V^a$ via trained $G$ as $\hat{Y}^a$.
    
    \noindent\jcst{\textbf{\textit{Stage II. Feature Encoder Fine-tuning.}}}
    \STATE Combine $E$ with self-guided attention module as $E_{SGA}$, then fine-tune $E_{SGA}$ with supervision of $Y^n \cup \hat{Y}^a$.
  \end{algorithmic}
\end{algorithm}
\vspace{-0.2cm}
\subsection{Overview}
\vspace{-0.2cm}
Given a video $V=\{v_i\}^N_{i=1}$ with $N$ clips, the annotated video-level label $Y\in\{1,0\}$ indicates whether an anomalous event exists in this video. 
We take a video $V$ as a bag and clips $v_i$ in the video as instances. Specifically, a negative bag (\ie $Y=0$) marked as $B^n = \{v^n_i\}^N_{i=1}$ has no anomalous instance, while a positive bag (\ie $Y=1$) denoted as $B^a = \{v^a_i\}^N_{i=1}$ has at least one. 

In this work, given a pair of bags (\ie a positive bag $B^a$ and a negative bag $B^n$), we first pre-extract the features (\ie $\{\mathbf{f}^a_i\}_{i=1}^N$ and $\{\mathbf{f}^n_i\}_{i=1}^N$ for $B^a$ and $B^n$, respectively) for each clip in the video $V=\{v_i\}^N_{i=1}$ using a pretrained vanilla feature encoder\jcst{, C3D or I3D, forming bags of features $\overline{B}^a$ and $\overline{B}^n$}. We then feed the pseudo label generator the extracted features to estimate the anomaly scores of the clips (\ie $\{s_i^a\}^N_{i=1}$, $\{s_i^n\}^N_{i=1}$).
Then, we produce pseudo labels $\hat{Y}^a=\{\hat{y}^a_i\}_{i=1}^N$ for anomalous video by performing smoothing and normalization on estimated scores to supervise the learning of the \ftnd{proposed self-guided attention boosted feature encoder}, \jcgam{forming as two-stage self-training scheme} \cite{lee2013pseudo,zou2018unsupervised,zou2019confidence}.

\ftnd{As shown in Figure \ref{fig:Strucutre}, our proposed feature encoder $E_{SGA}$, adapted from vanilla feature encoder $E$ (\eg, I3D or C3D) by adding our proposed self-guided attention module,} can be optimized with the estimated pseudo labels to eliminate the domain gap and produce task-specific representations. Actually, our proposed approach can be viewed as a two-stage method (see Algorithm \ref{alg:framework}): 1) we first generate clip-level pseudo labels for anomalous videos that have only video-level labels via the pseudo label generator, while the parameters of the pseudo label generator are updated by means of the deep MIL ranking loss. 2) After obtaining the clip-level pseudo labels of anomalous videos, our feature encoder $E_{SGA}$ can be trained on both normal and anomalous video data. Thus, we form a self-training scheme to optimize both the feature encoder $E_{SGA}$ and pseudo label generator $G$. The illustration shown in Figure \ref{fig:Strucutre} provides an overview of our proposed method.

To better distinguish anomalous clips from normal ones, we introduce a self-guided attention module in the feature encoder, \ie, $E_{SGA}$, to capture the anomalous regions in videos to help the feature encoder produce more discriminative representations (see Section \ref{sec:attention}). Moreover, we introduce a sparse continuous sampling strategy in the pseudo label generator to enforce
\jcnd{the network to pay more attention to the context around the most anomalous part}
(see Section \ref{sec:mil}). Finally, we introduce the deep MIL ranking loss to optimize the learning of the pseudo label generator, \ftst{and we use cross entropy loss to train our proposed feature encoder $E_{SGA}$ supervised by pseudo labels of anomalous videos and clip-level annotations of normal videos.}
\vspace{-0.1cm}
\subsection{Pseudo Label Generation via Multiple Instance Learning} \label{sec:mil}
\vspace{-0.1cm}
\begin{figure}[t]
    \centering
    \includegraphics[width=0.9\linewidth]{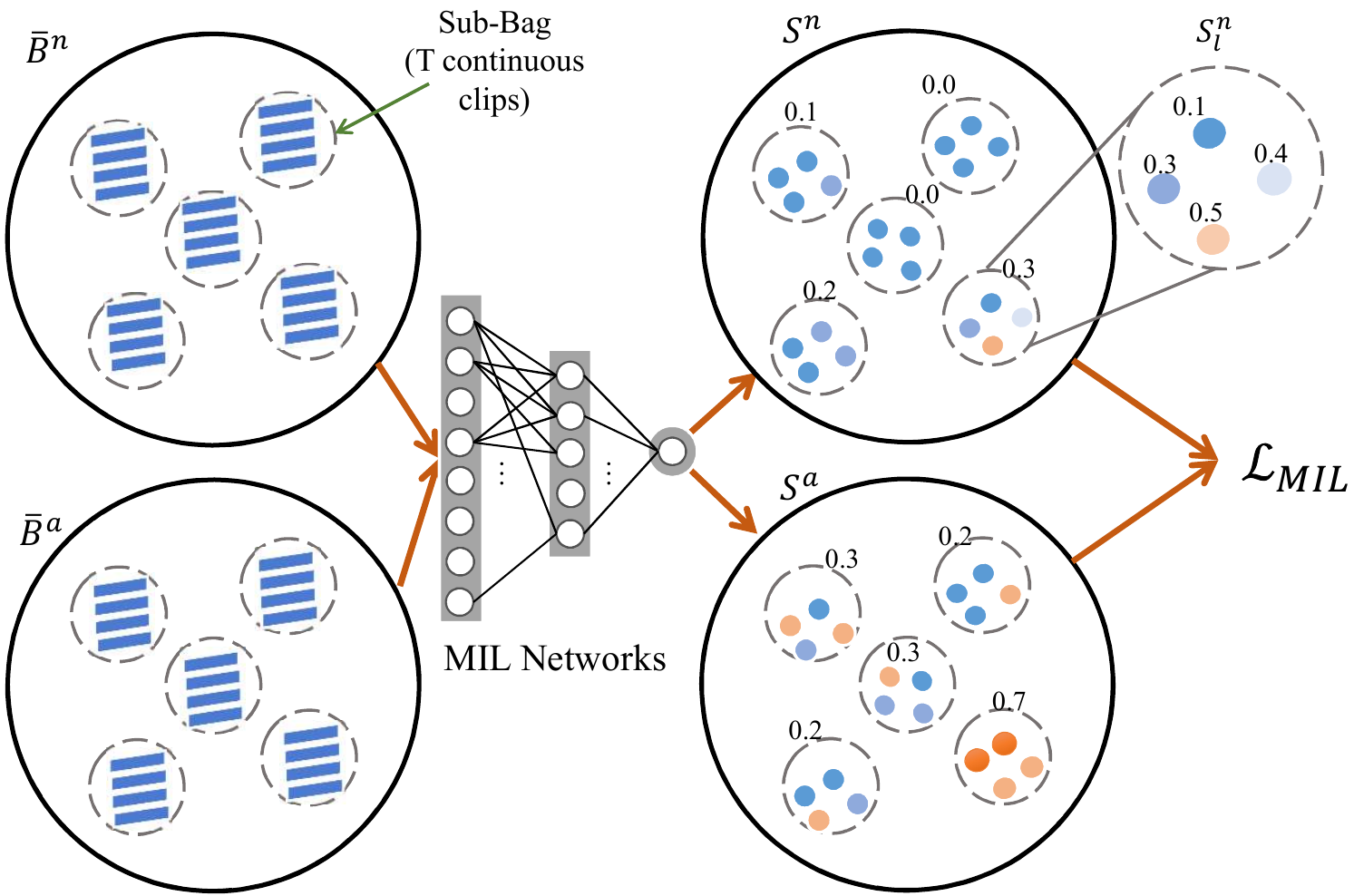}
    \vspace{-0.1cm}
    \caption{\ftnd{The workflow of our multiple instance pseudo label generator}. Each bag contains $L$ sub-bags, and each sub-bag is composed of $T$ continuous clips. }
    \label{fig:MIL_Flow}
    \vspace{-0.4cm}
\end{figure}
In contrast to \cite{zhong2019graph}, which simply assigns video-level labels to each clip and then trains the vanilla feature encoder at the very beginning, we introduce a \jcca{MLP-based} structure as the pseudo label generator \jcca{trained under the MIL paradigm} to generate pseudo labels, which are utilized in the refinement process of our feature encoder $E_{SGA}$.

Even though recent MIL-based methods \cite{sultani2018real,zhang2019temporal} have made considerable progress, the process of slicing a video into fixed segments \jcca{in an coarse-grained manner} regardless of its duration is prone to bury abnormal patterns as normal frames that usually constitute the majority, even in abnormal videos \cite{wan2020weakly}. 
\jcnd{However, by sampling with a smaller temporal scale \jcca{in a fine-grained manner}, the network may overemphasize on the most intense part of an anomaly but ignore the context around it. }
In reality, anomalous events often last for a while. With the assumption of minimum duration of anomalies, 
\jcnd{the MIL network is forced to pay more attention to the context around the most anomalous part.}

\jcnd{Moreover, to} adapt to the variation in duration of untrimmed videos and class imbalance \jcnd{in amount}, we introduce a sparse continuous sampling strategy: given the features for each clip extracted by a vanilla feature encoder $E$ from a video $\{\mathbf{f}_i\}^N_{i=1}$, we uniformly sample $L$ subsets from these video clips, and each subset contains $T$ consecutive clips, forming $L$ sub-bags $\overline{\mathcal{B}}=\{\mathbf{f}_{l,t}\}_{l=1,t=1}^{L,T}$, as shown in Figure \ref{fig:MIL_Flow}. 
\jcca{Remarkably, $T$, a hyperparameter to be tuned, also plays as the assumption of minimum duration of anomalies, as discussed in the previous paragraph.}
Here, we combine the MIL model 
with our continuous sampling strategy, as shown in Figure \ref{fig:MIL_Flow}. We feed extracted features into our pseudo label generator to produce corresponding anomalous scores $\{s_{l,t}\}_{l=1,t=1}^{L,T}$. Next, we perform average pooling of the predicted instance-level scores ${s}_{l,t}$ of each sub-bag score as $\mathcal{S}_l$ below, which can be utilized in Eq. \ref{equ:mil_loss}.

\begin{equation}
    \mathcal{S}_l=\frac{1}{T}\sum_{t=1}^{T}{s}_{l,t}.
\end{equation}

After training, the trained multiple instance pseudo label generator predicts clip-level scores for all abnormal videos marked as $S^a=\{s^a_i\}^N_{i=1}$.
By performing temporal smoothing with a moving average filter to relieve the jitter of anomaly scores with kernel size of $k$,
\begin{equation}
    \tilde{s}^a_i=\frac{1}{2k}\sum_{j=i-k}^{i+k} s^a_{j},
\end{equation}
and min-max normalization,
\begin{equation} \label{equ:pseudo_label}
    \hat{y}^a_i=\left (\tilde{s}^a_i-\min \tilde{S}^a\right )/(\max \tilde{S}^a-\min \tilde{S}^a)), i\in [1,N],
\end{equation}
we refine the anomaly scores into $\hat{Y}=\{\hat{y}^a_i\}_{i=1}^N$. Specifically, $\hat{y}^a_i$ is in $[0, 1]$ and acts as a soft pseudo label. Then, the pseudo labeled data $\{V^a, \hat{Y}^a\}$ are combined with clip-level labeled data $\{V^n, {Y}^n\}$ as $\{V, Y\}$ to fine-tune the proposed feature encoder $E_{SGA}$.

\vspace{-0.1cm}
\subsection{Self-Guided Attention in Feature Encoder}\label{sec:attention}
\vspace{-0.1cm}
\jcca{In contrast to} vanilla feature encoder $E$, which provides only task-agnostic representations for the down-stream task, we propose a self-guided attention boosted feature encoder $E_{SGA}$ adapted from $E$, which optimizes attention map generation \jcca{via} pseudo labels \jcca{supervision} to enhance the learning of task-specific representations.

\begin{figure}
    \centering
    \includegraphics[width=\linewidth]{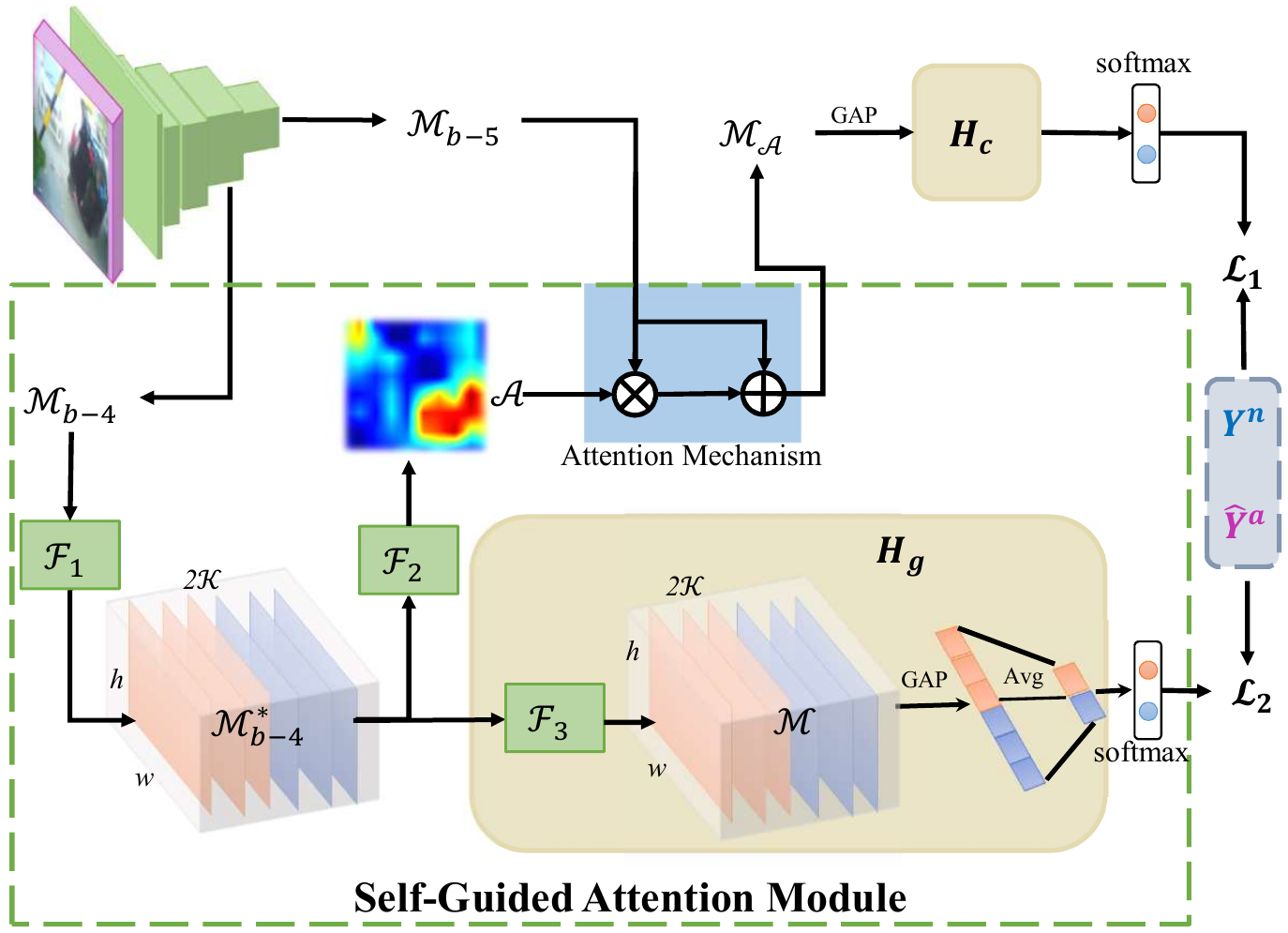}
    \caption{The \jcst{structure of self-guided attention boosted feature encoder $E_{SGA}$. GAP means the global average pooling operation, while Avg means $\mathcal{K}$ channel-wise average pooling in producing guided anomaly scores in guided classification head $H_g$. $\mathcal{A}$ is the attention map. \jcca{$\mathcal{F}_1$, $\mathcal{F}_2$, $\mathcal{F}_3$ are three encoding units constructed by convolutional layers.}}}
    \label{fig:attention}
    \vspace{-2mm}
\end{figure}

\ftst{As Figure \ref{fig:attention} shows, 
\jcst{the self-guided attention module (SGA) takes feature maps $\mathcal{M}_{b-4}$ and $\mathcal{M}_{b-5}$ as input, which are produced by the $4^{th}$ and $5^{th}$ blocks of vanilla feature encoder $E$, respectively. 
\jcca{SGA includes three encoding units, namely $\mathcal{F}_1$, $\mathcal{F}_2$ and $\mathcal{F}_3$, which are all constructed by convolutional layers.}
$\mathcal{M}_{b-4}$ is encoded as $\mathcal{M}^*_{b-4}$ and then applied to attention map $\mathcal{A}$ generation\jcca{,  denoted as}
\begin{equation}
    \mathcal{A}=\mathcal{F}_1(\mathcal{F}_2(\mathcal{M}_{b-4})).
\end{equation}
Finally, \ftst{we obtain} $\mathcal{M}_{A}$ via the attention mechanism below:}}
\begin{equation}
    \mathcal{M}_{\mathcal{A}}=\mathcal{M}_{b-5}+\mathcal{A} \circ \mathcal{M}_{b-5},
\end{equation}
\jcst{where \jcca{$\circ$ is element-wise multiplication, } and $\mathcal{M_A}$ is applied for final anomaly scores prediction via weighted-classification head $\mathcal{H}_c$, \ftnd{a fully connected layer.}}

\jcst{To assist the learning of the attention map, we introduce a guided-classification head $\mathcal{H}_g$ that uses the pseudo labels as supervision. \jcca{In} $\mathcal{H}_g$\jcca{, $\mathcal{F}_3$ transforms $\mathcal{M}^*_{b-4}$ into $\mathcal{M}$. Specifically, $\mathcal{M}^*_{b-4}$ and $\mathcal{M}$ hav $2\mathcal{K}$ channels as $\mathcal{K}$ multiple detectors for each class, \ie, normal and abnormal, to enhance the guided supervision \cite{yang2018weakly}.}
Then, we deploy spatiotemporal average pooling, $\mathcal{K}$ channel-wise average pooling on $\mathcal{M}$ and Softmax activation to obtain the guided anomaly scores for each class. 
}
%

\jcst{Remarkably, there are two classification heads in $E_{SGA}$, \ie, weighted-classification head $\mathcal{H}_c$ and guided classification head $\mathcal{H}_g$, which are both supervised by pseudo labels via $\mathcal{L}_1$ and $\mathcal{L}_2$, respectively. That is, we optimize  $E_{SGA}$ with the pseudo labels (see Section \ref{sec:mil}). Therefore, the feature encoder $E_{SGA}$ can update its parameters on video anomaly datasets and eliminate the domain gap from the pretrained parameters. }


\vspace{-0.1cm}
\subsection{Optimization Process}
\label{sec:obj_func}
\vspace{-0.1cm}

\noindent\textbf{- Deep MIL Ranking Loss:}
Considering that the positive bag contains at least one anomalous clip, we assume that the clip from a positive bag with the highest anomalous score is the most likely to be an anomaly \cite{hong2020mini}.
\jcnd{To adapt our sparse continuous sampling 
in \ref{sec:mil}, we treat a sub-bag as an instance and acquire a reliable relative comparison between the mostly likely anomalous sub-bag and the most likely normal sub-bag:
\begin{equation} \label{eq:reliable-ranking}
    \max_{\ftnd{1 \leq l \leq L}}{\mathcal{S}^n_l} <  \max_{\ftnd{1 \leq l \leq L}}{\mathcal{S}}^a_l
\end{equation}
Specifically, to avoid too many false positive instances in positive bags, we introduce a sparse constraint on positive bags, which instantiates Eq. \ref{eq:reliable-ranking} as a deep MIL ranking loss with sparse regularization:}
\vspace{0.05cm}
\begin{equation}\label{equ:mil_loss} 
\begin{aligned}
\mathcal{L}_{MIL} &=\left(\epsilon-\max_{1 \leq l \leq L}{\mathcal{S}^{a}_{l}}  + \max _{1 \leq l \leq L}{\mathcal{S}^{n}_{l}} \right)_{+} + \frac{\lambda}{L} \sum_{l=1}^{L} \mathcal{S}^{a}_{l}.
\end{aligned}
\end{equation}
where $(\cdot)_{+}$ means $\max(0,\cdot)$, and the first term in Eq. \ref{equ:mil_loss} ensures that $\max_{1 \leq l \leq L}{\mathcal{S}^a_l} $ is larger than $\max_{1 \leq l \leq L}{\mathcal{S}^n_l}$ with a margin of $\epsilon$. $\epsilon$ is a hyperparameter that is equal to 1 in this work. 
\jcgam{The last term in Eq. \ref{equ:mil_loss} is the sparse regularization indicating that only a few sub-bags may contain the anomaly,}
 while $\lambda$ is another hyperparameter used to balance the ranking loss with sparsity regularization.

\noindent\textbf{\ftnd{- Classification} Loss:} After obtaining the pseudo labels for an abnormal video in Eq. \ref{equ:pseudo_label}, we obtain the training pair $\{V^a, \hat{Y}^a\}$ that is further combined with $\{V^n, Y^n\}$ to train our feature encoder $E_{SGA}$. For this purpose, we apply the cross entropy loss function to the two classification heads ($\mathcal{H}_c$ and $\mathcal{H}_g$) in $E_{SGA}$, \ie $\mathcal{L}_{1}$ and $\mathcal{L}_{2}$ in Figure \ref{fig:attention}. 

\jcnd{Finally, we train a task-specific feature encoder $E_{SGA}$ with the combination of $\mathcal{L}_1$ and $\mathcal{L}_2$. In the inference stage, we use $E_{SGA}$ to predict clip-level scores for videos via weighted-classification head $\mathcal{H}_c$.}

\section{Experiments}
\subsection{Datasets and Metrics}
\vspace{-0.1cm}
We conduct experiments on two large datasets, \ie, UCF-Crime \cite{sultani2018real} and ShanghaiTech \cite{luo2017revisit}, with two feature encoders, \ie C3D \cite{tran2015learning} or I3D \cite{carreira2017quo}.

\noindent\textbf{UCF-Crime} is a large-scale dataset of real-world surveillance videos, including 13 types of anomalous events with 1900 long untrimmed videos, where 1610 videos are training videos and the others are test videos. Liu \et  \cite{liu2019exploring} manually annotated bounding boxes of anomalous regions in one image per 16 frames for each abnormal video, and we use their annotation of test videos only to evaluate our model's capacity to identify anomalous regions.

\noindent\textbf{ShanghaiTech} is a dataset of 437 campus surveillance videos. It has 130 abnormal events in 13 scenes, but all abnormal videos are in the test set, as the dataset is proposed for unsupervised learning. To adapt to the weakly supervised setting, Zhong \et  \cite{zhong2019graph} re-organized the videos into 238 training videos and 199 testing videos.

\noindent\jcca{\textbf{Evaluation Metrics.} Following previous works \cite{liu2018future,liu2019exploring,sultani2018real,wan2020weakly}, we compute the area under the curve (AUC) of the frame-level receiver operating characteristics (ROC) as the main metric, where a larger AUC implies higher distinguishing ability. We also follow \cite{sultani2018real,wan2020weakly} to evaluate robutness by the false alarm rate (FAR) of anomaly videos.}

\subsection{Implementation Details}
\vspace{-0.1cm}

The multiple instance pseudo label generator, is a 3-layer MLP, where the number of units  is 512, 32 and 1, respectively, regularized by dropout with probability of 0.6 between each layer. ReLU and Sigmoid functions are deployed after the first and last layer, respectively. Here, We adopt hyperparameters $L=32$, $T=3$, and $\lambda =0.01$ and train the generator with the \textit{Adagrad} optimizer with a learning rate of $0.01$. While \textit{fine-tuning}, we adopt the \textit{Adam} optimizer with a learning rate of $1e-4$ and a weight decay of 0.0005 and train 300 epochs. 
\ftnd{More details about implementation are reported in Supplementary Material.}
\subsection{Comparisons with Related Methods}
\vspace{-0.1cm}
\begin{figure}[t]
    \centering
    \includegraphics[width=0.9\linewidth]{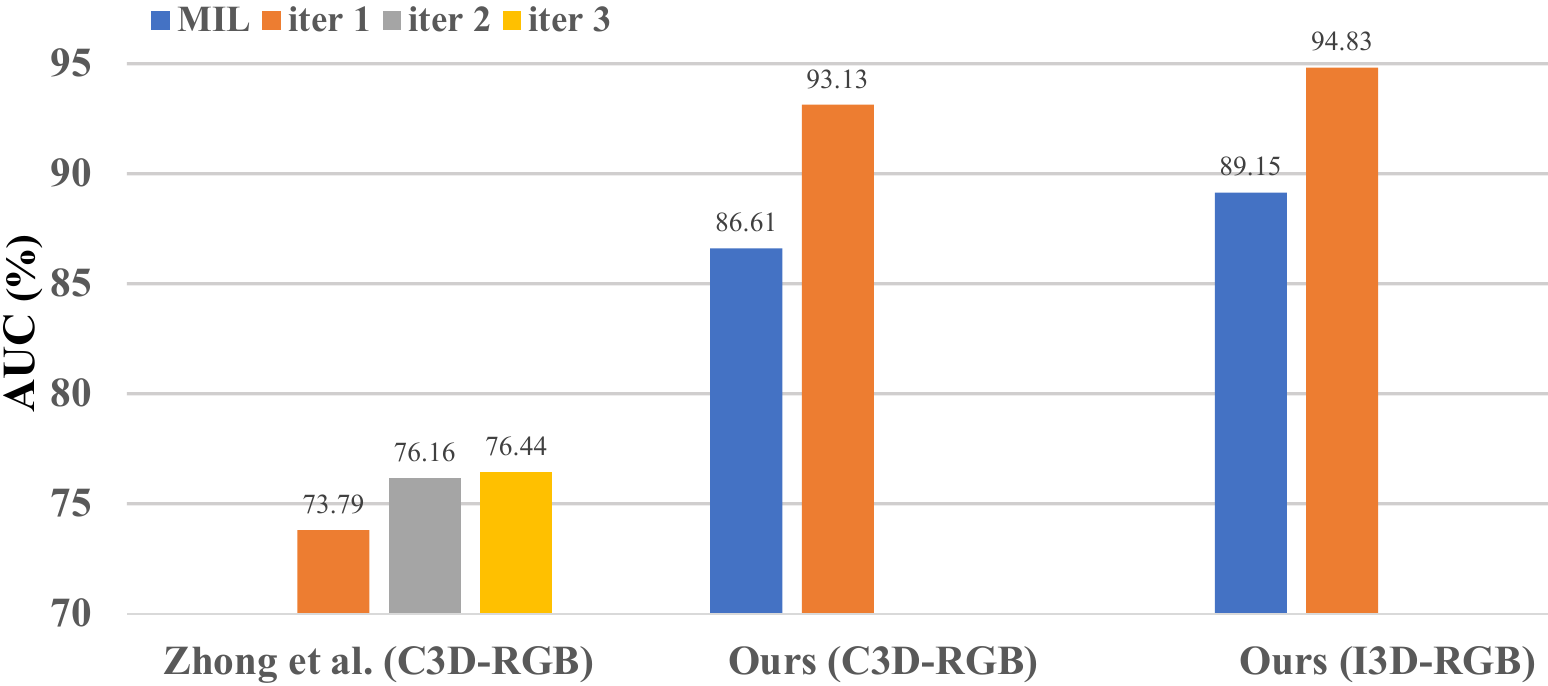}
    \vspace{-0.2cm}
    \caption{Comparisons with the state-of-the-art encoder-based method Zhong \et  \cite{zhong2019graph} on ShanghaiTech.}
    \vspace{-0.3cm}
    \label{fig:SHT_Chart}
\end{figure}

\begin{table}[t]
\center
\small
\scalebox{0.75}{
\begin{tabular}{cccccc}
\hline
Method                                 & Supervised & Grained  & Encoder                        & AUC (\%)  & FAR (\%) \\ \hline
\multicolumn{1}{c|}{Hasan et al. \cite{hasan2016learning}}      & \multicolumn{1}{c|}{Un} &\multicolumn{1}{c|}{Coarse}        & \multicolumn{1}{c|}{$AE^{RGB}$}   & 50.6         & 27.2 \\ 
\multicolumn{1}{c|}{Lu et al. \cite{lu2013abnormal}}      & \multicolumn{1}{c|}{Un}    &\multicolumn{1}{c|}{Coarse}    & \multicolumn{1}{c|}{Dictionary}   & 65.51         & 3.1 \\ \hline 
\multicolumn{1}{c|}{SVM}      & \multicolumn{1}{c|}{Weak} &\multicolumn{1}{c|}{Coarse} & \multicolumn{1}{c|}{$C3D^{RGB}$}   & 50        & - \\ 

\multicolumn{1}{c|}{Sultani et al. \cite{sultani2018real}}      & \multicolumn{1}{c|}{Weak} &\multicolumn{1}{c|}{Coarse} & \multicolumn{1}{c|}{$C3D^{RGB}$}   & 75.4         & 1.9 \\ 
\multicolumn{1}{c|}{Zhang et al. \cite{zhang2019temporal}}        & \multicolumn{1}{c|}{Weak} &\multicolumn{1}{c|}{Coarse}  & \multicolumn{1}{c|}{$C3D^{RGB}$}        & 78.7         & -    \\ 
\multicolumn{1}{c|}{Zhu et al. \cite{zhu2019motion}}          & \multicolumn{1}{c|}{Weak} &\multicolumn{1}{c|}{Coarse}  & \multicolumn{1}{c|}{$AE^{Flow}$}        & 79.0          & -      \\ 
\multicolumn{1}{c|}{Zhong et al. \cite{zhong2019graph}}        & \multicolumn{1}{c|}{Weak}&\multicolumn{1}{c|}{Fine}  &  \multicolumn{1}{c|}{$C3D^{RGB}$ }        & $80.67^*$(81.08)         & $3.3^*$(2.2)  \\ \hline  
\multicolumn{1}{c|}{Liu et al. \cite{liu2019exploring}}          & \multicolumn{1}{c|}{Full(T)} &\multicolumn{1}{c|}{Fine}   & \multicolumn{1}{c|}{$C3D^{RGB}$}& 70.1    & -      \\  
\multicolumn{1}{c|}{Liu et al. \cite{liu2019exploring}}          & \multicolumn{1}{c|}{Full(S+T)}&\multicolumn{1}{c|}{Fine}    & \multicolumn{1}{c|}{$NLN^{RGB}$}& 82.0    & -      \\ \hline 
\multicolumn{1}{c|}{MIST}          & \multicolumn{1}{c|}{Weak}&\multicolumn{1}{c|}{Fine} & \multicolumn{1}{c|}{$C3D^{RGB}$} & \textbf{81.40}          & 2.19     \\ 
\multicolumn{1}{c|}{MIST}          & \multicolumn{1}{c|}{Weak}&\multicolumn{1}{c|}{Fine} & \multicolumn{1}{c|}{$I3D^{RGB}$} & \textbf{82.30}         & \textbf{0.13}      \\  \hline 

\end{tabular}
}
\vspace{-0.1cm}
\caption{Quantitative comparisons with existing online methods on UCF-Crime \jcca{under different levels of supervision and fineness of prediction}. The results in $(\cdot)$ are tested with \textit{10-crop}, while those marked by ${}^*$ are tested without.}
\label{tab:comparision_ucf}
\vspace{-0.2cm}

\end{table}

In Table \ref{tab:comparision_ucf}, we present the \textbf{AUC}, \textbf{FAR} to compare our MIST with related state-of-the-art \jcca{online} methods in terms of \textit{accuracy and robustness}. We can find that MIST outperforms or performs similarly to all other methods in terms of all evaluation metrics from Table \ref{tab:comparision_ucf}, which confirms the efficacy of MIST.
Specifically, the results of Zhong \et \cite{zhong2019graph}, marked with $*$, are re-tested from the official released models\footnote{https://github.com/jx-zhong-for-academic-purpose/GCN-Anomaly-Detection.} without deploying \textit{10-crop}\footnote{10-crop is a test-time augmentation of cropping images into the center, four corners and their mirrored counterparts.} for fair comparison, while the results in brackets are reported on \cite{zhong2019graph} using 10-crop augmentation. However, 10-crop augmentation may improve the performance but requires $10$ times the computation. 
Notably, the result of our MIST still slightly overtakes that of Zhong \et  \cite{zhong2019graph} using 10-crop augmentation ($81.08\%$ vs. $81.40\%$ in terms of AUC and $2.2\%$ vs. $2.19\%$ for FAR).  Moreover, our method outperforms the supervised method of Liu \et  \cite{liu2019exploring}, which trains $C3D^{RGB}$ with external temporal annotations and $NLN^{RGB}$ with external spatiotemporal annotations. 
These results verify that our proposed MIST is more effective than previous works.

For the ShanghaiTech dataset results in Table \ref{tab:comparision_ShanghaiTech}, our MIST far outperforms other RGB-based methods \cite{sultani2018real,zhang2019temporal,zhong2019graph,wan2020weakly}, which validates the capacity of MIST. Remarkably, MIST also surpasses the multi-model method of AR-Net \cite{wan2020weakly} ($I3D^{RGB+Flow}$) on AUC by more than $4\%$ to $94.83\%$ and gains a much lower FAR of $0.05\%$.

\jcnd{We detail the comparison with the state-of-the-art encoder-based method \cite{zhong2019graph} on ShanghaiTech in Figure \ref{fig:SHT_Chart}. The multiple instance pseudo label generator performs much better than Zhong \et \ \cite{zhong2019graph}, which indicates the drawback of utilizing video-level labels as clip-level labels. Even though Zhong \et  \cite{zhong2019graph} optimizes for three iterations, it falls far behind our MIST with 16.69\% AUC on C3D, which solidly verifies the efficiency and efficacy of  MIST. Moreover, our MIST is much faster in the inference stage, as Zhong \et \ \cite{zhong2019graph} applies \textit{10-crop} augmentation.}


\begin{table}[t]
\center
\scalebox{0.82}{
\begin{tabular}{ccccc}
\hline
Method                      &  Feature Encoder    & Grained            & AUC (\%)          & FAR (\%)         \\ \hline
\multicolumn{1}{c|}{Sultani \et  \cite{sultani2018real}}    &  \multicolumn{1}{c|}{$C3D^{RGB}$}   & \multicolumn{1}{c|}{Coarse}               &  86.30           & 0.15           \\
\multicolumn{1}{c|}{Zhang \et  \cite{zhang2019temporal}}      &  \multicolumn{1}{c|}{$C3D^{RGB}$}   &  \multicolumn{1}{c|}{Coarse}               & 82.50           & 0.10           \\
\multicolumn{1}{c|}{Zhong \et  \cite{zhong2019graph}}      &  \multicolumn{1}{c|}{$C3D^{RGB}$}   & \multicolumn{1}{c|}{Fine}               & 76.44           & -              \\
\multicolumn{1}{c|}{AR-Net \cite{wan2020weakly}}           &  \multicolumn{1}{c|}{$C3D^{RGB}$}   & \multicolumn{1}{c|}{Fine}                & $85.01^*$           & $0.57^*$           \\
\multicolumn{1}{c|}{AR-Net \cite{wan2020weakly}}           &  \multicolumn{1}{c|}{$I3D^{RGB}$}  & \multicolumn{1}{c|}{Fine}                 & 85.38           & 0.27           \\
\multicolumn{1}{c|}{AR-Net \cite{wan2020weakly}}           &  \multicolumn{1}{c|}{$I3D^{RGB+Flow}$}  & \multicolumn{1}{c|}{Fine}                 & 91.24           & 0.10           \\
\hline
\multicolumn{1}{c|}{\textbf{MIST }}    &  \multicolumn{1}{c|}{$C3D^{RGB}$}   & \multicolumn{1}{c|}{Fine}               & \textbf{93.13}  & 1.71  \\
\multicolumn{1}{c|}{\textbf{MIST }}    &  \multicolumn{1}{c|}{$I3D^{RGB}$}   &\multicolumn{1}{c|}{Fine}               & \textbf{94.83}  & \textbf{0.05}  \\ \hline

\end{tabular}
}
\vspace{-0.1cm}
\caption{Quantitative comparisons with existing methods on ShanghaiTech. The results with ${}^*$ are re-implemented.}
\label{tab:comparision_ShanghaiTech}
\vspace{-0.2cm}
\end{table}

\begin{table}[t]
\scalebox{0.8}{
\begin{tabular}{c|cccc}
\hline
\multirow{3}{*}{\begin{tabular}[c]{@{}c@{}}Encoder-Agnostic\\  Methods\end{tabular}} & \multicolumn{4}{c}{AUC (\%)}                                                                       \\ \cline{2-5} 
                                                                                     & \multicolumn{2}{c|}{UCF-Crime}                     & \multicolumn{2}{c}{ShanghaiTech} \\
                                                                                     & pretrained  & \multicolumn{1}{c|}{fine-tuned} & pretrained   & fine-tuned   \\ \hline
Sultani \et  \cite{sultani2018real}                                                   & 78.43              & \multicolumn{1}{c|}{81.42}                 & 86.92             &   92.63
\\
Zhang \et  \cite{zhang2019temporal}                                                   & 78.11              & \multicolumn{1}{c|}{81.58}        & 88.87            &   92.50                \\
AR-Net \cite{wan2020weakly}                                                      & 78.96        & \multicolumn{1}{c|}{82.62}            & 85.38             &   92.27        \\
Our MIL generator                                                                    & 79.37             & \multicolumn{1}{c|}{81.55}            &       89.15             &       92.24        \\ \hline
\end{tabular}
}
\vspace{-0.1cm}
\caption{Quantitative comparisons between the features from the pretrained vanilla feature encoder and those from MIST on UCF-Crime and ShanghaiTech datasets by adopting encoder-agnostic methods.}
\label{tab:encoder_evaluate}
\vspace{-0.2cm}
\end{table}

\begin{figure}
    \centering
    \scalebox{0.85}{
    \includegraphics[width=\linewidth]{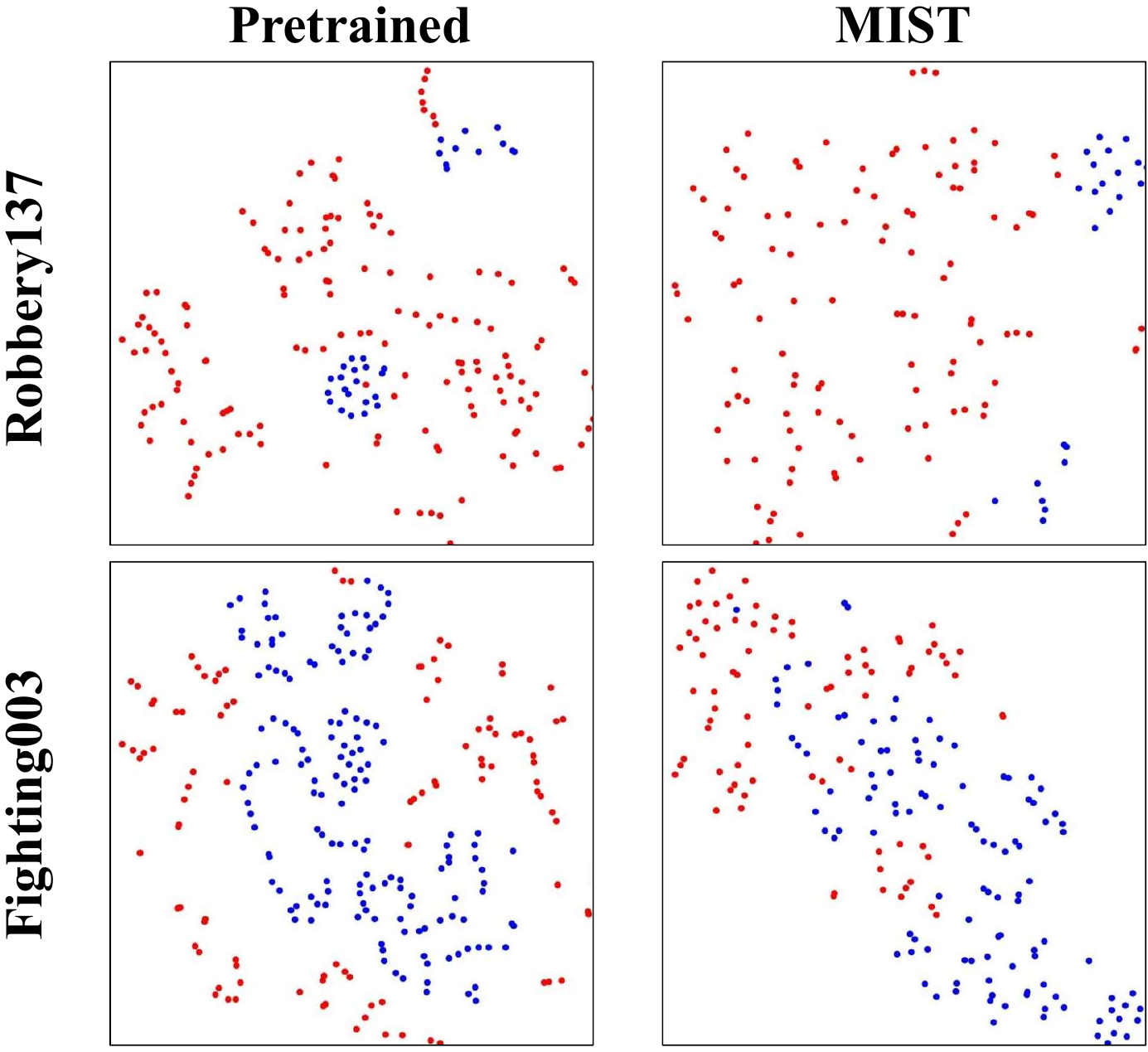}
    }
    \caption{Feature space visualization of pretrained vanilla feature encoder \textbf{I3D} and the MIST fine-tuned encoder via t-SNE \cite{maaten2008visualizing} on UCF-Crime testing videos. The red dots denote anomalous regions while the blue ones are normal. }
    \vspace{-2mm}
    \label{fig:tsne_I3D}
\end{figure}

\vspace{-0.1cm}
\subsection{Task-Specific Feature Encoder}
\vspace{-0.2cm}
\ftst{To verify that \ftnd{our} feature encoder can produce task-specific representations that facilitate the other encoder-agnostic methods, we also conduct related experiments with  \jcst{\textbf{I3D}} as presented in Table \ref{tab:encoder_evaluate}.
It is noticeable that all results of encoder-agnostic methods are boosted after using our MIST fine-tuned features,
\jcgam{showing a reduction in the domain gap}. 
For example, AR-Net \cite{wan2020weakly} increases from 85.38\% to 92.27\% on the UCF-Crime dataset and achieves an improvement of 6.89\% on the ShanghaiTech dataset. Therefore, our MIST can produce a more powerful \ftnd{task-specific} feature encoder that can be utilized in other approaches. We visualize the feature space of the pretrained I3D vanilla feature encoder and the MIST-fine-tuned encoder via t-SNE\cite{maaten2008visualizing} in Figure \ref{fig:tsne_I3D}, which also indicates the refinement of feature representations.}

   

\begin{table}[t]
\scalebox{0.8}{
    \begin{tabular}{c|c|cc|c}
    \hline
    \multirow{2}{*}{Dataset} & \multirow{2}{*}{Feature} & \multicolumn{2}{c|}{AUC (\%)}                        & \multirow{2}{*}{\begin{tabular}[c]{@{}c@{}}$\Delta$AUC\\(\%)  \end{tabular}} \\ \cline{3-4}
                      &                   & \multicolumn{1}{c|}{Uniform} & Sparse Continuous &                   \\ \hline \hline
    \multirow{2}{*}{UCF-Crime} & $C3D^{RGB}$& \multicolumn{1}{c|}{74.29}  &    75.51     &          +1.22         \\ \cline{2-5} 
                      &           $I3D^{RGB}$        & \multicolumn{1}{c|}{78.72}                &  79.37       &       +0.65            \\ \hline
    \multirow{2}{*}{ShanghaiTech} & $C3D^{RGB}$& \multicolumn{1}{c|}{83.68}                &      86.61    &        +2.93          \\ \cline{2-5} 
                      &           $I3D^{RGB}$        & \multicolumn{1}{c|}{83.10}               &  89.15       &        +6.05           \\ \hline
    \end{tabular}
}
\caption{\jcca{Performance comparisons of sparse continuous sampling and uniform sampling for MIL generator training.} }
\label{tab:sampling}
\end{table}

            

\vspace{-0.1cm}
\subsection{Ablation Study}
\vspace{-0.1cm}
At first, we introduce another evaluation metric, \ie \emph{score gap}, which is the gap between the average scores of abnormal clips and normal clips. Larger score gap indicates \jcnd{the network is more capable of distinguishing anomalies from \ftnd{normal} events }\cite{liu2018future}. 
\jcnd{We conduct ablation studies on UCF-Crime to analyze the impact of generated pseudo labels (PLs), the self-guided attention module (SGA), and classifier head $H_g$ in SGA of proposed feature encoder $E_{SGA}$ in Table \ref{tab:ablation_ucf}. Compared with the baseline and MIST$^{\textup{w/o PLs}}$, our MIST achievees a significant improvement when the generated pseudo labels are utilized. In particular, we observe 8.17\% improvement in AUC and an approximately 17\% score gap, which shows the efficacy of our multiple instance pseudo label generator with the sparse continuous sampling strategy. }\jcca{Pseudo labels also plays an important role. Compared with MIST, the performance of MIST$^{\textup{w/o PLs}}$ drops seriously, even worse than the baseline for the low-quality supervision that influences the attention map $\mathcal{A}$ generation from SGA. }

\jcgam{Moreover, SGA enhances the feature encoder on emphasizing the informative regions and distinguishing abnormal events from normal ones. Compared with $MIST^{w/o SGA}$, MIST increases by 2\% in AUC and 5\% in the score gap. Specifically, the guided-classification branch in SGA plays an important role in guiding the attention map generation, and there is a drop of more than 2\% if such a branch is removed. }


\jcca{Ablation studies are also conducted on a sparse continuous sampling strategy on UCF-Crime and ShanghaiTech with $C3D^{RGB}$ and $I3D^{RGB}$ features. As shown in Table \ref{tab:sampling}, when sampling the same number of clips for a bag and selecting the same number of top clips to represent the bag, our sparse continuous sampling strategy pays more attention to the context and does better than uniform sampling. Especially in ShanghaiTech, sparse continuous sampling gains 2.93\% and 6.05\% on two kinds of features.}




\begin{table}
\center
\scalebox{0.9}{
\begin{tabular}{c|cc}
\hline
Method & AUC (\%)& Score Gap (\%) \\ 

 \hline
Baseline& 74.13 & 0.375  \\
MIST$^{\textup{w/o PLs}}$ &  73.33 & 0.443 \\
MIST$^{\textup{w/o } H_g}$  &  81.97 & 15.37  \\ 
MIST$^{\textup{w/o SGA}}$&  80.28 & 12.74  \\\hline
MIST  &  \textbf{82.30} & \textbf{17.71}  \\ \hline

\end{tabular}
}
\vspace{-0.1cm}
\caption{Ablation Studies on UCF-Crime with $I3D^{RGB}$. Baseline is the original \textbf{I3D} trained with video-level labels \cite{zhong2019graph}. MIST is our whole model. $\textup{MIST}^\textup{w/o PLs}$ is trained without pseudo labels but with video-level labels. $\textup{MIST}^{\textup{w/o }H_g}$ is MIST trained without $H_g$. $\textup{MIST}^{\textup{w/o SGA}}$ is trained without the self-guided attention module).}
\label{tab:ablation_ucf}
\vspace{-0.2cm}
\end{table}

\begin{figure*}[ht]
    \centering
    \vspace{-0.5cm}
    \includegraphics[width=\textwidth]{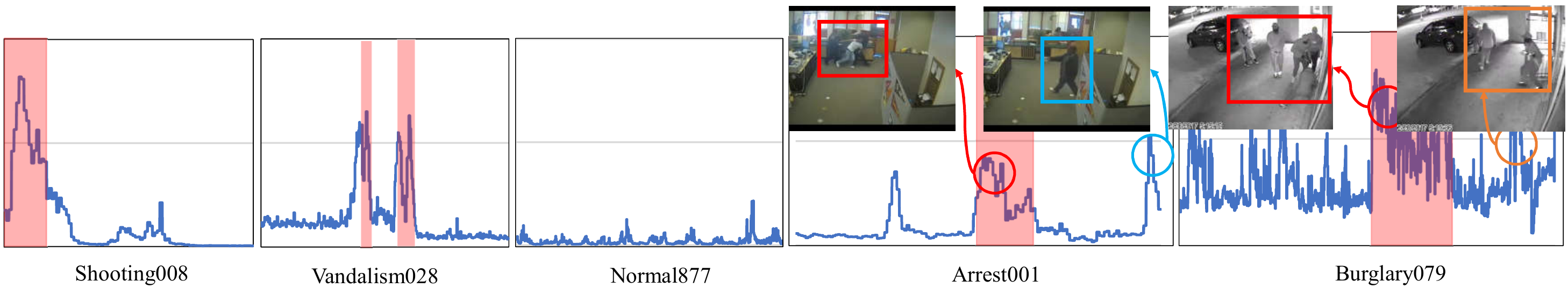}
    \vspace{-0.7cm}
    \caption{Visualization of the testing results on UCF-Crime (better viewed in color). The red blocks in the graphs are temporal ground truths of anomalous events. The orange circle shows the wrongly labeled ground truth, the blue circle indicates the wrongly predicted clip, and the red cricle indicates the correctly predicted clip. }
    \label{fig:ano_scores}
    \vspace{-0.3cm}
\end{figure*}

\begin{figure}[ht]
    \centering
    \scalebox{1}{
    \vspace{-0.2cm}
    \includegraphics[height=5cm,width=\linewidth]{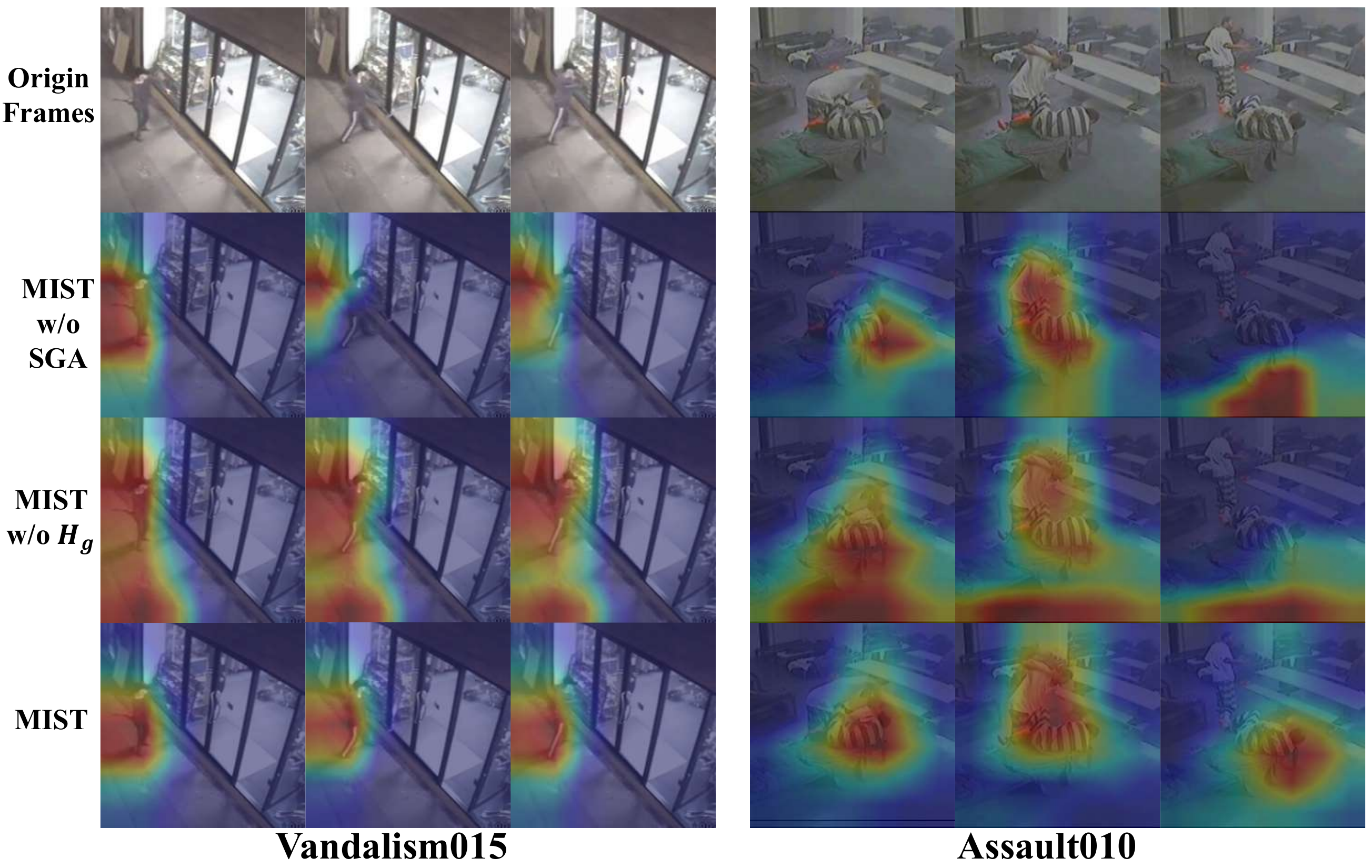}
    }
    \vspace{-0.4cm}
    \caption{Visualization results of anomaly activation maps (better viewed in color). }
    \vspace{-0.3cm}
    \label{fig:spa_vis}
\end{figure}

\vspace{-0.1cm}
\subsection{Visual Results}
\vspace{-0.1cm}
To further evaluate the performance of our model, we visualize the temporal predictions of the models. As presented in Figure \ref{fig:ano_scores}, our model exactly localizes the anomalous events and predicts anomaly scores very close to zero on normal videos, showing the effectiveness and robustness of our model. We collect some failed samples in the right row of Figure \ref{fig:ano_scores}. In addition, our model predicts the highest score at the end of \textit{Arrest001}, where a man walks across the scene with his arm pointing forward as if brandishing a gun. \ftst{As the videos in UCF-Crime are low-resolution, it is difficult to judge such a confusing action without any other context information.}
Furthermore, the bottom-right part of Figure \ref{fig:ano_scores}
 shows another failed case; \ie, our model successfully localizes the major part of the anomalous burglary event and raises an alarm when the thieves are rushing out of the house, which should be treated as an anomaly but is wrongly labeled as a normal event in the ground truth.
We also visualize the spatial activation map via Grad-CAM \jcca{on $\mathcal{M}_{\mathcal{A}}$} \cite{selvaraju2017grad} for spatial explanation. As Figure \ref{fig:spa_vis} shows, our model is able to sensitively focus on informative regions that help decide whether the scene is anomalous . \ftst{This verifies that our self-guided attention module can boost the feature encoder to focus on anomalous regions. }\jcgam{Additionally, compared with the activation maps generated from the MIST without guided-classification head $H_g$ and the MIST without the SGA module, the results of MIST are concentrated on the anomalous regions, which shows the rationality and effectiveness of our self-guided attention module.}
\vspace{-0.1cm}

\subsection{Discussions}
\vspace{-0.1cm}
The key of our MIST is to design a two stage self-training strategy to train a task-specific feature encoder for video anomaly detection. Each component of our framework can be replaced by any other advanced module, \eg, replacing C3D with I3D, or a stronger pseudo label generator to take the place of the multiple instance pseudo label generator. Additionally, the scheme of our framework can be adapted to other tasks, such as weakly supervised video action localization and video highlight detection. 

\vspace{-0.1cm}
\section{Conclusions}
\vspace{-0.1cm}
\jcgam{In this work, we propose a multiple instance self-training framework (MIST) to fine-tune a task-specific feature encoder efficiently.}
We adopt a sparse continuous sampling strategy in the multiple instance pseudo label generator to produce more reliable
pseudo labels. \jcnd{With the estimated pseudo labels, 
 our proposed feature encoder} learns to focus on the most probable anomalous regions in frames facilitated by the proposed self-guided attention module. 
\jcgam{Finally, after a two-stage self-training process, we train a task-feature encoder with discriminative representations that can also boost other existing methods. Remarkably, our MIST makes significant improvements on two public datasets.}

\section*{Acknowledgement}
\noindent\jcca{This work was supported partially by the National Key Research and Development Program of China (2018YFB1004903), NSFC (U1911401, U1811461), Guangdong Province Science and Technology Innovation Leading Talents (2016TX03X157), Guangdong NSF Project (Nos.2020B1515120085, 2018B030312002), Guangzhou Research Project (201902010037), Research Projects of Zhejiang Lab (No.2019KD0AB03), and the Key-Area Research and Development Program of Guangzhou (202007030004).}

\clearpage
\newpage
\section*{\LARGE{Appendix}}\label{s:appendix}
\appendix
\section{Comparisons of Action Recognition Datasets and Anomaly Detection Datasets}
\label{sec:diff}
\begin{figure}
    \centering
    \includegraphics[width=\linewidth]{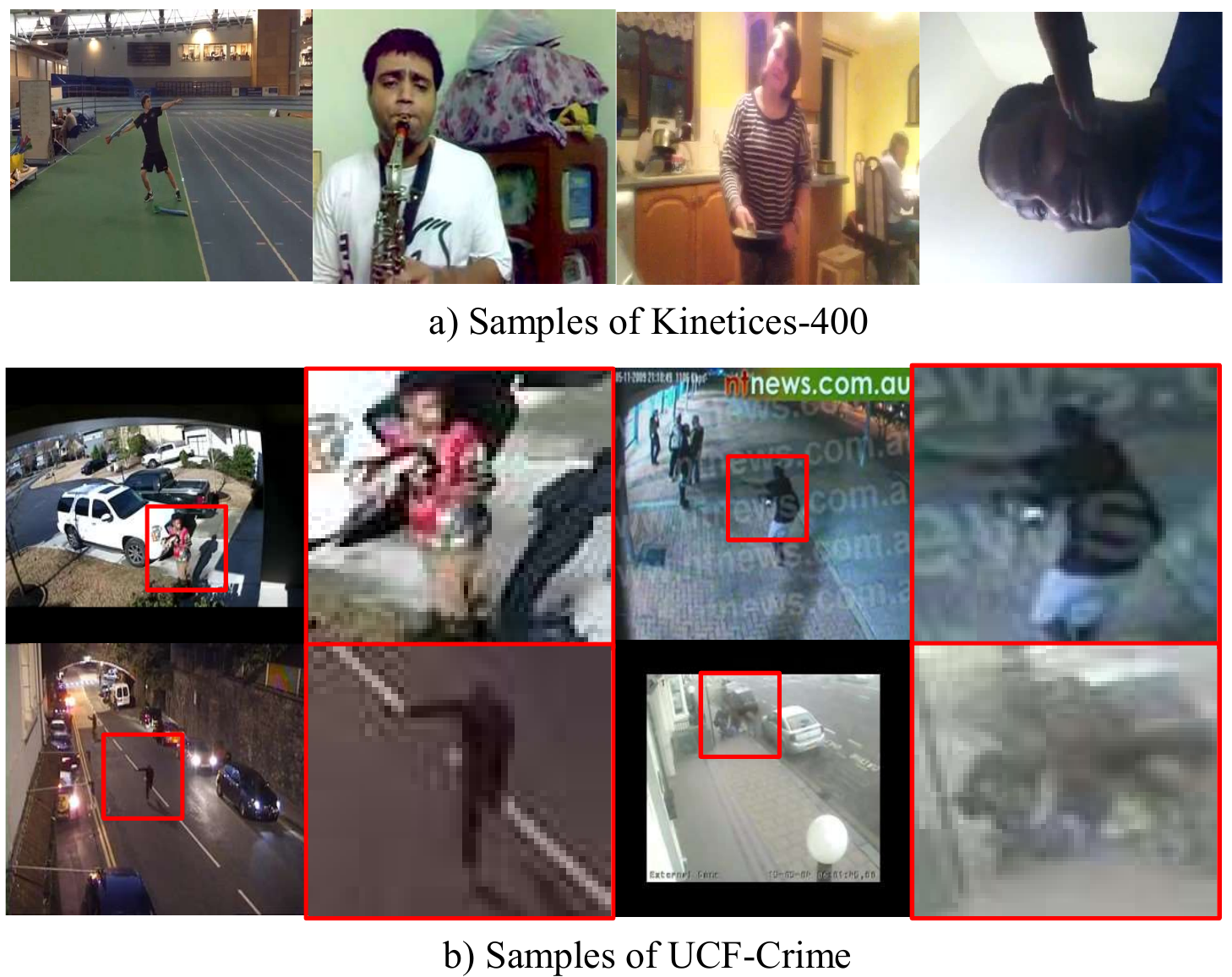}
    \caption{Samples of action recognition dataset Kinetics-400 \cite{kay2017kinetics} and video anomaly dataset UCF-Crime \cite{sultani2018real}. The red boxes are the anomalous regions in frames and their corresponding enlarged images.}
    \label{fig:dataset_compare}
\end{figure}
As Figure \ref{fig:dataset_compare} shown, the samples from Kinetics-400  \cite{kay2017kinetics} are actor-centered while the samples from UCF-Crime \cite{sultani2018real} are not \cite{choi2019can,choi2020unsupervised}. Additionally, the anomalies in frames are usually small and low-resolution. These situations indicate the domain gap between the two kinds of datasets. In this work, we propose MIST to minimize the domain gap by training both feature encoder and classifier in a two stage self-training scheme. 

\section{Details of Pseudo Label Generation}
\subsection{Feature Extraction and Sampling}
\begin{figure}
    \centering
    \includegraphics[width=\linewidth]{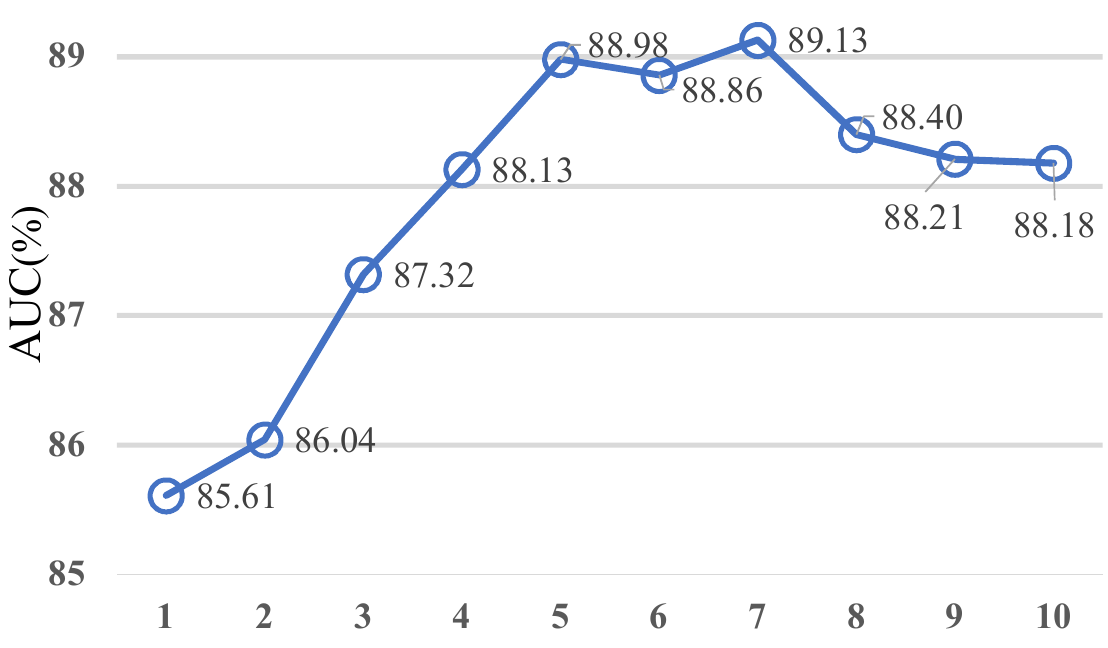}
    \caption{The effect of $T$ for a fixed number of sub-bag 32 on ShanghaiTech dataset with $I3D^{RGB}$ features.}
    \label{fig:SHT_param}
\end{figure}
We deploy a vanilla feature encoder, \ie \textbf{C3D} \cite{tran2015learning} pretrained on Sport-1M \cite{karpathy2014large} or \textbf{I3D} pretrained on Kinetics-400 \cite{kay2017kinetics} to extract features for generator training. We densely sample 16 frames per clip most of the times but 12 frames per clip for \textbf{I3D} on UCF-Crime. After extracting the features, sparse continuous sampling is applied to sample the $L \cdot T$ clips to form bags of features $\overline{B}$. Then, $\mathcal{L}_{MIL}$ is deployed to optimize the generator. 
Specifically, we follow \cite{sultani2018real} to select $L=32$. As for $T$, we choose $T=3$ for UCF-Crime and $T=7$ for ShanghaiTech. We have shown the selection of $\mathcal{K}$ on ShanghaiTech with $I3D^{RGB}$ features in Figure \ref{fig:SHT_param}. Additionally, $\lambda$ is set as 0.01. 40 normal and 40 abnormal videos are randomly sampled as a batch when training. 

\subsection{Pseudo Label Refinement}
The trained generator predicts clip-level scores $S^a=\{s^a_i\}^N_{i=1}$ for all abnormal videos in the training set. Temporal moving average filter with kernel size $k=5$ and min-max normalization are deployed to refine the anomaly scores into $\hat{Y}=\{\hat{y}_i^a\}_{i=1}^N$.

\section{Details of Feature Encoder Finetuning}
\begin{figure}
    \centering
    \includegraphics[width=\linewidth]{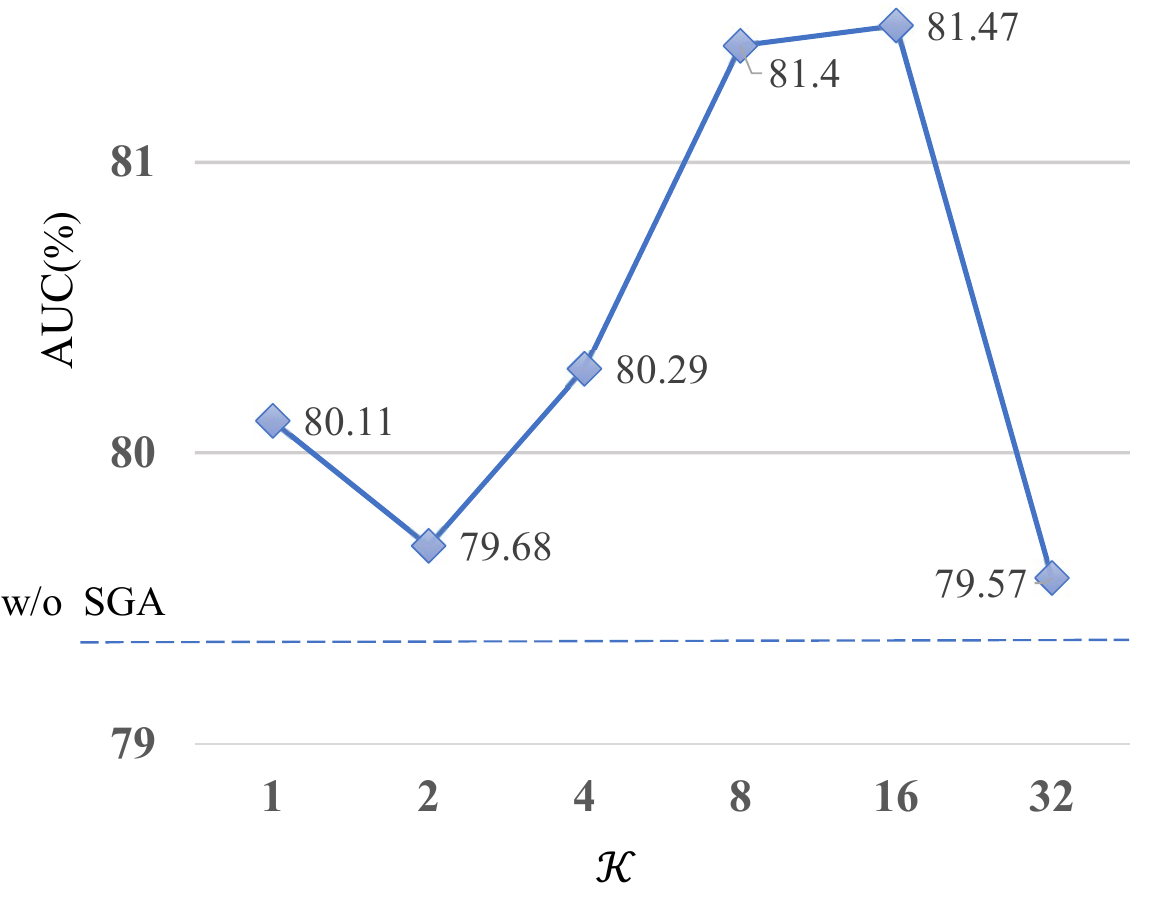}
    \caption{Variations of AUC for different values of multiple detector $\mathcal{K}$ with \textbf{C3D} on UCF-Crime dataset. The dotted line is the result of MIST training without self-guided attention module.}
    \label{fig:k_select}
\end{figure}

\subsection{\jcca{Implementation of Self-Guided Attention Module}}
\jcca{As shown in Figure 4 of the submission, our proposed self-guided attention module includes 3 encoding units, namely $\mathcal{F}_1$, $\mathcal{F}_2$, $\mathcal{F}_3$. All of these encoding units are constructed by convolutional layers. Let $C$ represents the number of channel of $\mathcal{M}_{b-4}$. $\mathcal{F}_1$ consists of a $3\times3\times3\times C$ 3DConv layer with the stride of 2 and a $1\times1\times1\times2\mathcal{K}$ 3DConv layer, which are both activated by ReLU function; $\mathcal{F}_2$  is a $1\times1\times1\times1$ 3DConv layer activated by Sigmoid function; $\mathcal{F}_3$ is a $1\times1\times1\times 2\mathcal{K}$ 3DConv layer. Then, the attention map $\mathcal{A}$ is calculated as follows: 
\begin{equation}
    \mathcal{A}=\mathcal{F}_2(\mathcal{F}_1(\mathcal{M}_{b-4})),
\end{equation}
while the guided classification prediction $\hat{p}$ is an aggregation results from $\mathcal{M}$, which is calculated below:
\begin{equation}
    \mathcal{M}=\mathcal{F}_3(\mathcal{F}_1(\mathcal{M}_{b-4})).
\end{equation}
Specifically, $\hat{p}$ is transformed from $\mathcal{M}$ via spatiaotemporal average pooling $\Pi$ and class-specific channel-wise average pooling $\Phi$:
\begin{equation}
    \hat{p}=\Phi(\Pi(\mathcal{M})),
\end{equation}
which is further optimized by $\mathcal{L}_2$ to guide the optimization of class-wise discriminative feature map $\mathcal{M}_{b-4}^*$ and then strengthen the attention map generation indirectly.}

\subsection{\jcca{Implementation of $E_{SGA}$ Finetuning}}
For UCF-Crime, we sample 16 abnormal videos and 16 normal videos per batch, and uniformly sample 3 clips from each video. For ShanghaiTech, we sample 10 abnormal videos and 10 normal videos per batch. The training process finishes in 300 epochs. Specifically, at the begining of finetuning, we conduct \textit{warm-up} for 5 epochs. Since only a few clips of the abnormal video are anomalous, there exists a class-imbalance problem, especially for \textbf{I3D}. We introduce class-reweighting to cross-entropy loss as class-weighted cross-entropy loss $\mathcal{L}_{w}$:
\begin{equation}
    \mathcal{L}_{w}=-w_0y\textup{log}p-w_1(1-y)\textup{log}(1-p),
\end{equation}
where $w_0$ and $w_1$ are class weights for abnormal and normal class, respectively. \jcca{Specifically, $\mathcal{L}_1$ and $\mathcal{L}_2$ are adopted the same kind of loss function $\mathcal{L}_w$} We adopt $w_0=1.2$ and $w_1=0.8$ for UCF-Crime, while $w_0=0.8$ and $w_1=0.65$ for ShanghaiTech. 

In the left of Figure \ref{fig:k_select}, we report the AUC of STSA with different $\mathcal{K}$. The performance goes up as the $\mathcal{K}$ get larger and reaches the top with $\mathcal{K}$ of 8 or 16. When the value getting even larger, it seems to be overfitting and get worse. Considering a trade-off between the efficiency with effectiveness, we set $\mathcal{K}=8$ in our framework for all other experiments.  

\begin{figure*}
    \centering
    \includegraphics[width=\textwidth]{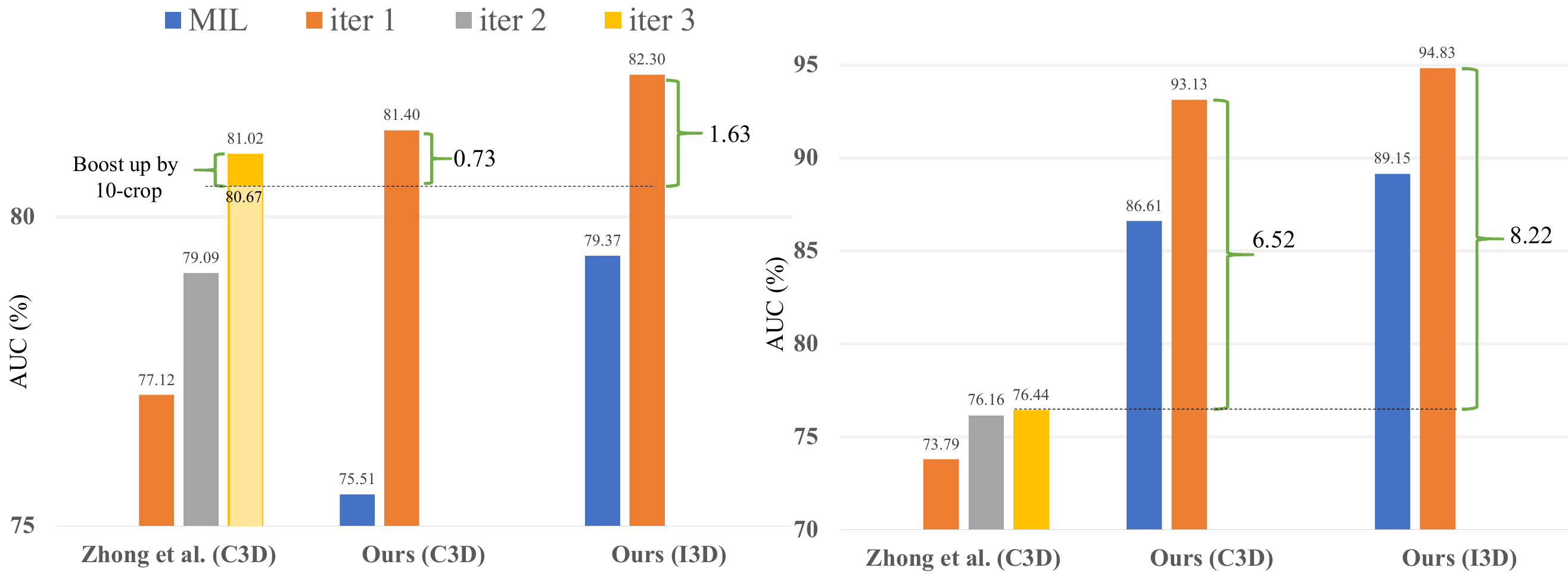}
    \caption{Quantitative Comparisons with state-of-the-art encoder-based method Zhong \et \cite{zhong2019graph} on UCF-Crime and ShanghaiTech.}
    \label{fig:zhong_compare}
\end{figure*}

\begin{figure*}
    \centering
    \includegraphics[width=\textwidth]{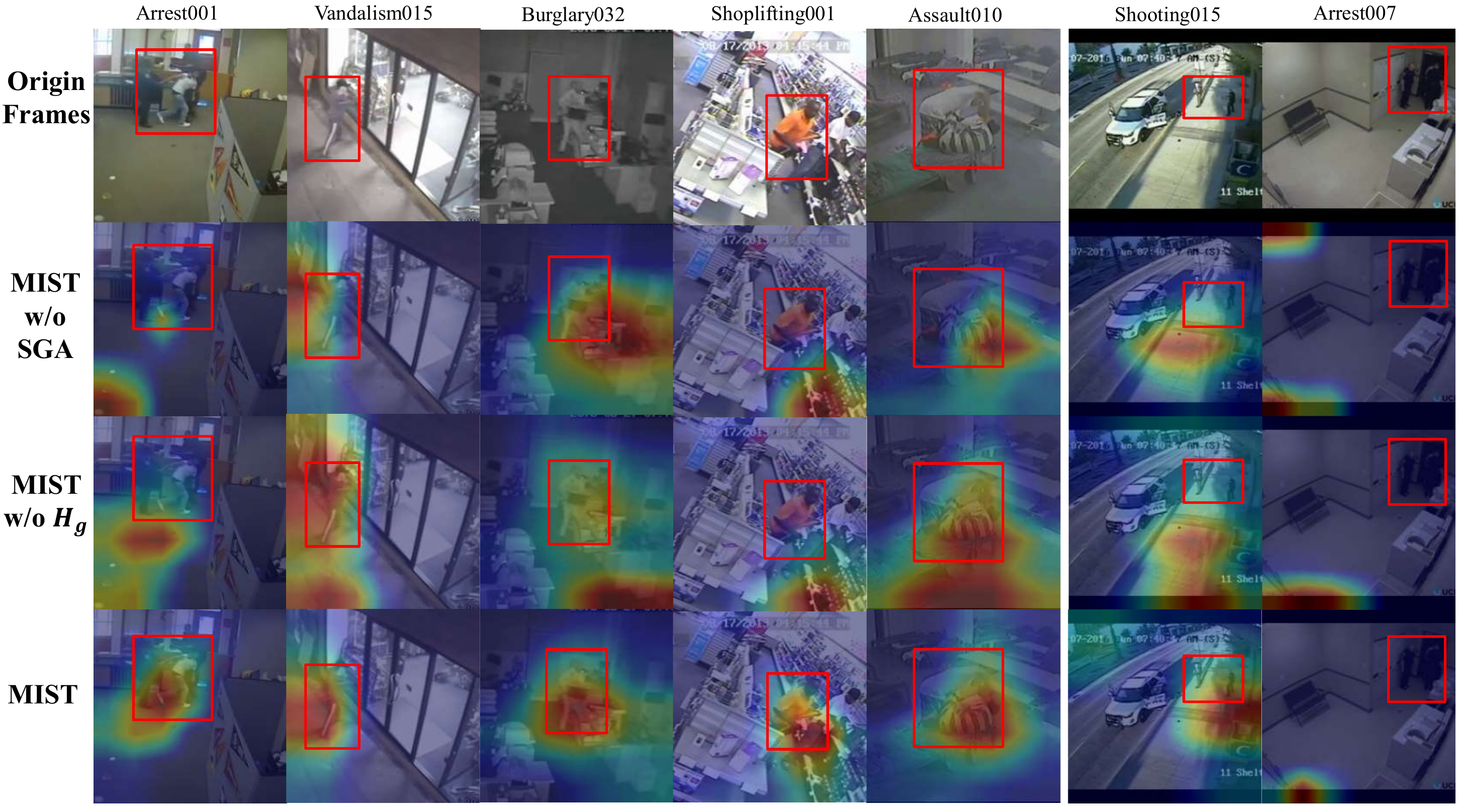}
    \caption{More spatial anomaly activation maps visualization on UCF-Crime. The left 5 columns of the graphs are the successful results while the right 2 columns are the failures. The red boxes are the ground-truth spatial annotations \cite{liu2019exploring}.}
    \label{fig:grad_cam}
\end{figure*}

After finetuning, we acquire a task-specific feature encoder $E_{SGA}$. $E_{SGA}$ outperforms state-of-the-art encoder-based method Zhong \et \ \cite{zhong2019graph}, which is shown in Figure \ref{fig:zhong_compare} in detail. Moreover, $E_{SGA}$ can focus on the anomalous regions in frames, which is shown in Figure \ref{fig:grad_cam}. As the left 5 columns of the figure shown, self-guided attention module help the feature encoder in focusing the anomalous regions. We have also listed the failure on the right 2 columns of the figure, the results from too small size of anomaly regions.

\section{\jcca{More Experimental Results}}
\subsection{\jcca{Speed and Computational Complexity}}
\begin{table}[h]
    \centering
    \scalebox{0.85}{
        \begin{tabular}{c||c|c|c}
        \hline
            Model & \#Params& Speed (FPS) & FLOPs (MAC)\\ \hline
             MIST-I3D & 31M & 324.46 & 45.68G \\ \hline
             MIST-C3D & 85M & 197.10  & 39.26G \\ \hline
             Zhong-C3D\cite{zhong2019graph} & 78M  & 130.04  & 386.2G\\ \hline
        \end{tabular}
    }
    \caption{Speed and computational complexity comparisons with Zhong \et \cite{zhong2019graph}.}
    \label{tab:speed}
\end{table}

\jcca{There are four 1080Ti GPUs for stage 2 but one 1080Ti GPU for stage 1 and validation.
In C3D (I3D) based model, \#Params are 85 M (31 M), the FLOPs are 39.26 G (45.68 G) and the speed is 197.10 FPS (324.46 FPS). Compared to Zhong \et that adopt \textit{10-crop} testing time augmentation, our method is much faster but costs much lower computational complexity as shown in Table \ref{tab:speed}.}

\subsection{\jcca{More Quantitative Comparisons}}
\jcca{We show more quantitative comparisons with Zhong \et \cite{zhong2019graph} on UCF-Crime and ShanghaiTech on Figure \ref{fig:zhong_compare}. We observe a huge improvement in ShanghaiTech. As for UCF-Crime our method still do much better when compared fairly without using \textit{10-Crop}. Moreover, our method does much better on iter 1 as MIST does not need iterative optimization.}

\subsection{\jcca{More Spatial Visualization}}
\jcca{We also present more spatial visualization in Figure \ref{fig:grad_cam}. We observe that MIST performs better than those without SGA or $H_g$. The left two columns are the failure case where the front ground is extremely small and vague to be detected.}

\section{Discussions of the Formulation}
\subsection{ Label Noise Learning vs MIST}
Zhong \et \ \cite{zhong2019graph} treats weakly supervised video anomaly detection as a label noise learning task. However, the extreme label noise results from assigning video-level labels to each clip. In contrast, MIST offers pseudo labels with lower noise via multiple instance generator, which is more efficient. Additionally, MIST can further co-operate with label noise learning methods to refine pseudo labels iteratively and train a more powerful feature encoder.

\begin{table}[h]
    \centering
    \begin{tabular}{c|c|c||c}
         \hline
         Model & Before (\%) & After (\%) & Gain(\%)\\
         \hline \hline
         MIST-C3D & 58.66 & 67.14& +8.48\\ \hline
         MIST-I3D & 63.63 & 73.37& +9.74\\ \hline
    \end{tabular}
    \caption{Performance comparisons of before and after refinement on ShanghaiTech in term of AUC scores of anomaly videos. }
    \label{tab:refine}
\end{table}

\jcca{In contrast to Zhong \et that reduces the noise via a specific module, \ie GCN-based label noise cleaner, we resist label noise via post procession  likes min-max norm and temporal smoothing. As shown in Table \ref{tab:refine}, we conduct these two types of refinement are do a great help in removing label noise. Moreover, we also use large a batch size with the help of gradient accumulation to reduce the label noise \cite{rolnick2017deep}.}

\subsection{2D Feature Encoder vs 3D Feature Encoder}
We also conduct experiment on 2D feature encoder the RGB branch of TSN \cite{wang2018temporal} but fail. Similar result is also reported in \cite{liu2019exploring}. Since the RGB branch of TSN operates only on a single frame, it fails in catching the motion to represent temporal information. Instead, we deploy two popular 3D spatiotemporal feature encoders, \ie \textbf{C3D} and \textbf{I3D}, whose results well-verified the capacity of MIST.

\subsection{\jcca{Fine-Grained vs Coarse-Grained and Online vs Offline}}
\jcca{Our method focuses on online fine-grained anomaly detection. Previous works follow Sultani \et \cite{sultani2018real} to perform anomaly detection in a  coarse-grained manner. However, in the real world, we expect anomaly detection can be applied for streaming surveillance videos to detect anomalies precisely and quickly, while the methods in coarse-grained do not meet the requirement. Some work like Ullah \et \cite{ullah2020cnn} performs anomaly detection in an offline manner based on an external assumption as complete observation of the testing videos. As discussed above, it also 
violates the expectation for detection on streaming video.  }

{\small
\bibliographystyle{ieee_fullname}
\bibliography{MIST_for_Arxiv}
}

\end{document}